\pdfoutput=1

\documentclass[11pt]{article}

\usepackage{acl}

\usepackage{times}
\usepackage{latexsym}

\usepackage[T1]{fontenc}

\usepackage[utf8]{inputenc}
\usepackage{bm}
\usepackage{amsmath,amssymb}
\usepackage{booktabs}

\usepackage{microtype}

\usepackage{xspace}
\usepackage{enumitem}
\usepackage{tikz-dependency}
\usepackage{pifont}
\usepackage{CJKutf8}
\usepackage{multirow}
\usepackage{caption}
\usepackage{subcaption}
\title{Substructure Distribution Projection \\
for Zero-Shot Cross-Lingual Dependency Parsing}

\author{Haoyue Shi \qquad \qquad Kevin Gimpel \qquad \qquad Karen Livescu \\
  Toyota Technological Institute at Chicago \\
  6045 S Kenwood Ave, Chicago, IL, USA, 60637  \\
  \texttt{\{freda, kgimpel, klivescu\}@ttic.edu} \\}

\newcommand{\ddp}{\textsc{SubDP}\xspace} 
\newcommand{\freda}[1]{\textcolor{orange}{[Freda: #1]}}

\newcommand{\interalia}[1]{\citep[][\emph{inter alia}]{#1}}

\begin{document}
\begin{CJK*}{UTF8}{bsmi}

\maketitle
\begin{abstract}
We present substructure distribution projection (\ddp), a technique that projects a distribution over structures in one domain to another, by projecting substructure distributions separately. 
Models for the target domains can be then trained, using the projected distributions as soft silver labels. 
We evaluate \ddp on zero-shot cross-lingual dependency parsing, taking dependency arcs as substructures: we project the predicted dependency arc distributions in the source language(s) to target language(s), and train a target language parser to fit the resulting distributions. 
When an English treebank is the only annotation that involves human effort, \ddp achieves better unlabeled attachment score than all prior work on the Universal Dependencies v2.2 \citep{nivre-etal-2020-universal} test set across eight diverse target languages, as well as the best labeled attachment score on six out of eight languages. 
In addition, \ddp improves zero-shot cross-lingual dependency parsing with very few (e.g., 50) supervised bitext pairs, across a broader range of target languages.

\end{abstract}

\section{Introduction}
\begin{figure}[t!]
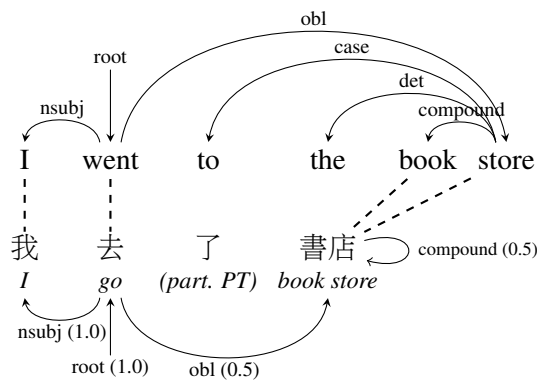
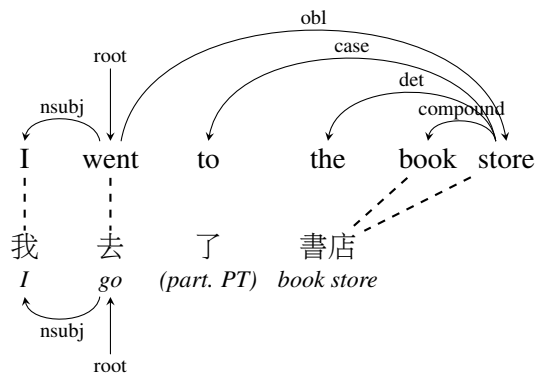

    \begin{subfigure}[b]{0.45\textwidth}
    \centering
    \begin{dependency}[arc edge, arc angle=80, text only label, label style={above}]
    \begin{deptext}[column sep=.1cm]
    I \& went \& to \& the \& book \&store \\[0.8cm]
    ~~~我~~~ \& 去 \& 了 \& 書店 \\
    \small \textit{I} \& \small \textit{go} \& \small \textit{(part. PT)} \& \small \textit{book store}\\
    \end{deptext} 
    \depedge{2}{1}{nsubj}
    \depedge{2}{6}{obl}
    \depedge{6}{5}{compound}
    \depedge{6}{3}{case}
    \depedge{6}{4}{det}
    \deproot[edge unit distance=0.3cm]{2}{root}
    \draw [-, thick, dashed] (\wordref{1}{1}) -- (\wordref{2}{1});
    \draw [-, thick, dashed] (\wordref{1}{2}) -- (\wordref{2}{2});
    \draw [-, thick, dashed] (\wordref{1}{5}) -- (\wordref{2}{4});
    \draw [-, thick, dashed] (\wordref{1}{6}) -- (\wordref{2}{4});
    \depedge[label style=below,edge below]{2}{1}{nsubj (1.0)}
    \depedge[label style=below, edge below]{2}{4}{obl (0.5)}
    \deproot[label style=below,edge below,edge unit distance=0.25cm]{2}{root (1.0)}
    \path (\wordref{2}{4}) edge [loop right] node {\scriptsize \textcolor{black}{compound (0.5)}} (\wordref{2}{4});
    \end{dependency}  \\[-0.2cm]
    \flushright {\small part. PT: particle denoting past tense.} \\[0.2cm]
    \caption{Dependency tree projection with \ddp.}
    \label{fig:teaser-soft}
    \end{subfigure}
    \begin{subfigure}[b]{0.45\textwidth}
    \centering
    \begin{dependency}[arc edge, arc angle=80, text only label, label style={above}]
    \begin{deptext}[column sep=.1cm]
    I \& went \& to \& the \& book \&store \\[0.8cm]
    ~~~我~~~ \& 去 \& 了 \& 書店 \\
    \small \textit{I} \& \small \textit{go} \& \small \textit{(part. PT)} \& \small \textit{book store}\\
    \end{deptext} 
    \depedge{2}{1}{nsubj}
    \depedge{2}{6}{obl}
    \depedge{6}{5}{compound}
    \depedge{6}{3}{case}
    \depedge{6}{4}{det}
    \deproot[edge unit distance=0.3cm]{2}{root}
    \draw [-, thick, dashed] (\wordref{1}{1}) -- (\wordref{2}{1});
    \draw [-, thick, dashed] (\wordref{1}{2}) -- (\wordref{2}{2});
    \draw [-, thick, dashed] (\wordref{1}{5}) -- (\wordref{2}{4});
    \draw [-, thick, dashed] (\wordref{1}{6}) -- (\wordref{2}{4});
    \depedge[label style=below,edge below]{2}{1}{nsubj}
    \deproot[label style=below,edge below,edge unit distance=0.25cm]{2}{root}
    \end{dependency}
    \caption{Projection with only one-to-one alignments.}
    \label{fig:teaser-hard}
    \end{subfigure}
    \caption{Illustration of \ddp (top) vs.\ a representative of annotation projection \citep[bottom;][]{lacroix-etal-2016-frustratingly}. 
    An English parse tree, labeled with the Universal Dependencies conventions \citep{nivre-etal-2016-universal,nivre-etal-2020-universal}, is projected to the parallel sentence in Chinese. We denote dependency edges by arrows with the corresponding arc probabilities in parentheses, and word alignments by dashed lines. 
    } 
    \label{fig:teaser}
\end{figure}

Zero-shot cross-lingual dependency parsing has benefited from cross-lingual grounding signals such as bitext \interalia{ma-xia-2014-unsupervised,lacroix-etal-2016-frustratingly,rasooli-etal-2021-wikily}.
A popular line of work is annotation projection: the parses generated by a source language dependency parser are projected into the target language, where the projected parses are then used to train a new parser. 
As illustrated in Figure~\ref{fig:teaser-hard}, most annotation projection methods 
typically output partial dependency parse trees, where there either is or is not an arc between any pair of words. 
In addition, most bitext-based 
work has relied on one-to-one word alignment between bitext pairs \citep[e.g., \textit{I} and~\textit{我}~in Figure~\ref{fig:teaser};][\emph{inter alia}]{ma-xia-2014-unsupervised,lacroix-etal-2016-frustratingly,rasooli-etal-2021-wikily}, discarding many-to-one alignments (e.g., \textit{book store} and~\textit{書店}~in Figure~\ref{fig:teaser}). 

In this work, we introduce substructure distribution projection (\ddp), where dependency arcs act as substructures. 
We consider projecting substructure distributions, i.e., the conditional probability distribution of the corresponding head given a word, instead of the distribution of whole parse trees \citep{ma-xia-2014-unsupervised}.
Taking the case where the source is a discrete parse tree 
(Figure~\ref{fig:teaser-soft}) as an example, \ddp has the same behavior as prior work   \citep[e.g.,][]{lacroix-etal-2016-frustratingly} for arcs that are only involved in one-to-one alignments; for many-to-one alignments, \ddp projects the corresponding arcs into \emph{soft} arc distributions in the target language. 
Therefore in \ddp, a target language word may have multiple heads in the projected trees, where their probabilities sum to one. 

More generally, \ddp may take dependency arc or label distributions in the source language(s), instead of discrete trees, as the input. 
Following the annotation projection schema \interalia{rasooli-collins-2015-density,lacroix-etal-2016-frustratingly,zhang-etal-2019-cross}, the projected soft trees are then used to train a target language dependency parser. 

We mainly evaluate \ddp on zero-shot cross-lingual dependency parsing with eight selected languages from the Universal Dependencies v2.2 \citep{nivre-etal-2020-universal}, where the English treebank is the only supervision that involves human effort. 
Taking English as the source language, \ddp significantly outperforms all baseline methods on all distant languages in our experiments
(Arabic, Hindi, Korean and Turkish), in terms of both labeled attachment scores (LAS) and unlabeled attachment scores (UAS), while achieving superior UAS on all nearby languages (German, French, Spanish and Italian) as well. 
Further analysis shows that \ddp also helps improve zero-shot cross-lingual dependency parsing with a small amount of supervised bitext, across a broader range of target languages. 

\section{Related Work}
\label{sec:related}
\paragraph{Zero-shot cross-lingual dependency parsing.\footnote{Also referred to as zero-shot dependency parsing in recent literature \citep{schuster-etal-2019-cross,wang-etal-2019-cross}.}}
Zero-shot cross-lingual dependency parsing is a task that requires a model to predict
dependency parses without seeing any parsing example in the target language; instead, the model may use annotated parses in other languages.  
Existing approaches can be classified into the following categories: 
\begin{enumerate}[leftmargin=*,itemsep=0mm]
    \item \textbf{Delexicalized training} \interalia{zeman-resnik-2008-cross,mcdonald-etal-2011-multi,cohen-etal-2011-unsupervised,durrett-etal-2012-syntactic,rosa-zabokrtsky-2015-klcpos3}, which only considers delexicalized features (e.g., part-of-speech tags) in the training process, and seeks to transfer to the target languages.  
    \item \textbf{Transfer with cross-lingual embeddings} \interalia{tackstrom-etal-2012-cross,guo-etal-2015-cross,schuster-etal-2019-cross}, which assumes that cross-lingual word representations, including word clusters \citep{tackstrom-etal-2012-cross,ammar-etal-2016-many}, word type embeddings \citep{guo-etal-2015-cross,guo2016representation,duong-etal-2015-cross,ammar-etal-2016-many,wick2016minimally}, and contextualized cross-lingual word embeddings \citep{schuster-etal-2019-cross,wang-etal-2019-cross,he-etal-2019-cross,ahmad-etal-2019-difficulties,ahmad-etal-2019-cross}, provide shared features for words with similar syntactic roles. 
    \item \textbf{Treebank translation} \citep{tiedemann-etal-2014-treebank,tiedemann-2015-improving}, which seeks to translate treebanks in the source language(s) into the target languages or a code-switching mode \citep{zhang-etal-2019-cross}, and use the translated treebanks to train target language parsers. 
    \item \textbf{Annotation projection}, which trains a parser in the source language(s), and projects the predicted source language parse trees to target languages using bitext \citep{hwa2005bootstrapping,ma-xia-2014-unsupervised,agic-etal-2016-multilingual}.  
    Additional strategies are usually used to improve the projection quality, such as keeping confident edges only \citep{li-etal-2014-soft,lacroix-etal-2016-frustratingly}, projection from multiple source languages \citep{agic-etal-2016-multilingual,rasooli-collins-2017-cross}, density based iterative filtering \citep{rasooli-collins-2015-density,rasooli-collins-2017-cross,rasooli-collins-2019-low} and noisy self-training \citep{kurniawan-etal-2021-ppt}. 
\end{enumerate}

There are different assumptions on annotation availability, such as gold part-of-speech tags  \interalia{zeman-resnik-2008-cross,cohen-etal-2011-unsupervised,durrett-etal-2012-syntactic}, 
a reasonably good translator, which uses extra annotated data in the training process \citep{tiedemann-etal-2014-treebank,tiedemann-2015-improving,zhang-etal-2019-cross}, 
high-quality bilingual lexicons \interalia{durrett-etal-2012-syntactic,guo-etal-2015-cross,guo2016representation}, 
and language-specific constraints \citep{meng-etal-2019-target}. 
Most bitext-based work assumes annotated bitext 
\interalia{ma-xia-2014-unsupervised,li-etal-2014-soft,lacroix-etal-2016-frustratingly} or constructed bitext from extra signals \citep[e.g., Wikipedia; ][]{rasooli-etal-2021-wikily}. However,  \citet{he-etal-2019-cross}, \citet{schuster-etal-2019-cross}, \citet{ahmad-etal-2019-difficulties,ahmad-etal-2019-cross} and \citet{kurniawan-etal-2021-ppt} only require minimal annotations (i.e., source language treebanks and unlimited raw text in relevant languages). 

In this work, we propose a distribution projection method to project dependency arc and label distributions for zero-shot cross-lingual dependency parsing, falling into the category of annotation projection. 
Some of the benefits of our method relative to prior work are that it works well with minimal annotations, allows soft word alignment (\S\ref{sec:preliminary}), supports both labeled and unlabeled parsing, and has a low time complexity $\mathcal{O}(n^2)$ for non-projective parsing.\footnote{In contrast, \citet{ma-xia-2014-unsupervised} requires $\mathcal{O}(n^4)$ time for non-projective unlabeled dependency parsing.} 

\paragraph{Multilingual contextualized representations.} 
Recent contextualized models pretrained on multilingual text \interalia{devlin-etal-2019-bert,conneau-etal-2020-unsupervised,tran-etal-2020-cross} have been demonstrated to be effective across a wide range of cross-lingual NLP tasks, including bitext retrieval \citep{tran-etal-2020-cross}, cross-lingual named entity recognition \citep{pires-etal-2019-multilingual,mulcaire-etal-2019-polyglot}, and cross-lingual dependency parsing \citep{schuster-etal-2019-cross,wang-etal-2019-cross}. 
In this work, we apply two of the contextualized pretrained models, XLM-R \citep{conneau-etal-2020-unsupervised} and CRISS \citep{tran-etal-2020-cross} to generate unsupervised bitext. 

\paragraph{Soft-label methods.} 
As an intuitive extension to the regular cross entropy loss against one-hot labels, calculating the cross entropy between model output and a soft distribution has been applied to knowledge distillation \interalia{hinton-etal-2015-distilling,you2017learning,sanh2019distillbert}, cross-lingual named entity recognition \citep{wu-etal-2020-single} and to handle annotation discrepancy \citep{fornaciari-etal-2021-beyond}. 
Our approach is an instantiation of soft-label methods on cross-lingual transfer learning, with additional post processing to the output of the original models. 
\section{Proposed Approach: \ddp}
We start this section with the introduction to the background (\S\ref{sec:background}) and preliminaries (\S\ref{sec:preliminary}). 
Our proposed pipeline for zero-shot cross-lingual dependency parsing consists of three steps: 
(1) training a bi-affine dependency parser $\mathcal{P}_1$ in the source language $L_1$, 
(2) projecting annotations on $L_1$ sentences to their corresponding parallel sentences in the target language $L_2$ (\S\ref{sec:ddp}), and 
(3) training another bi-affine dependency parser $\mathcal{P}_2$ 
for $L_2$ (\S\ref{sec:optimization}). 

\subsection{Background}
\label{sec:background} 
\paragraph{Bi-affine dependency parser.} 
For a sentence with $n$ words $\langle w_1, \ldots, w_n\rangle$,\footnote{For convenience, we assume that $w_1$ is an added dummy word that has one dependent -- the root word of the sentence.} we denote the word features when acting as heads and dependents by $\bm{H}\in \mathbb{R}^{n \times d_h}$ and $\bm{D} \in \mathbb{R}^{n \times d_d}$ respectively, where $d_h$ and $d_d$ denote the dimensionality of the corresponding features. 
The probability of word $w_i$ having the head $w_j$ can be formulated as an $n$-way classification problem:
\begin{align}
    \bm{S}^\textit{(arc)} &= \bm{D}\bm{W}^{\textit{(arc)}}\bm{H}^\intercal \label{eq:bi-affine} \\
    P(w_j \mid w_i) &= \frac{\exp\left(\bm S^\textit{(arc)}_{i, j}\right)}{\sum_{k=1}^{|s|} \exp\left(\bm S^\textit{(arc)}_{i, k}\right)}, \label{eq:arc-prob}
\end{align}
where $\bm W^{\textit{(arc)}} \in \mathbb{R}^{d_d\times d_h}$ is the parameters of the bi-affine module.\footnote{While Eq~\eqref{eq:bi-affine} is in a bi-linear form, in practice, we can always append a constant feature column to both $\bm H$ and $\bm D$, resulting in a bi-affine model.}  
Given $\log P(w_j \mid w_i)$ for every pair of $i$ and $j$, the dependency trees can be 
inferred by finding the spanning arborescence of maximum weight using the Chu–Liu-Edmonds algorithm \citep{chu1965shortest,edmonds1968optimum}. 
We use the algorithm proposed by \citet{tarjan1977finding}, which has an $\mathcal{O}(n^2)$ time complexity for each individual sentence. 

As for dependency arc labeling, we denote the candidate label set by $L$. 
Parameterized by $\bm{W}^\textit{(label)} \in \mathbb{R}^{d_d\times d_h \times |L|}$, we define the probability that the arc from head $s_j$ to dependent $s_i$ 
has the label $\ell$ by 
\begin{align}
    \bm{S}^\textit{(label)}_{i, j, \ell} &= \sum_{p} \sum_{q} \bm{D}_{i,p}\bm{W}^{(label)}_{p, q, \ell} \bm{H}_{j,q} \nonumber \\
    P(\ell\mid w_j\rightarrow w_i) &= \frac{\exp\left(\bm S^\textit{(label)}_{i, j, \ell}\right)}{\sum_{k=1}^{|L|} \exp\left(\bm S^\textit{(label)}_{i, j, k}\right)}, \label{eq:label-prob}
\end{align}
Given the probability definitions above, the model is trained to maximize the log likelihood of the training data. 
More details can be found in \citet{dozat2016deep}.

We use bi-affine dependency parsers as the backbone for all parsers in this work, while it is worth noting that \ddp works for any parser that produces a set of arc and label distributions. 

\paragraph{CRISS}
CRISS \citep{tran-etal-2020-cross} is an unsupervised machine translation model trained with monolingual corpora, starting from mBART \citep{liu-etal-2020-mbart}, a multilingual pretrained sequence-to-sequence model with a mask-filling denoising objective.
During the training process, CRISS iteratively (1) encodes sentences in the monolingual corpora with its encoder, (2) mines bitext based on encoding similarity, and (3) uses the mined bitext to fine-tune the model with a machine translation objective. 
In this work, we use CRISS to generate unsupervised translation of English sentences to construct bitext, and apply its encoder to extract word features for an ablation study. 

\paragraph{SimAlign} SimAlign \citep{jalili-sabet-etal-2020-simalign} is a similarity based word aligner: given a pair of source and target sentence $\langle s, t\rangle$, SimAlign computes a contextualized representation for each token in both $s$ and $t$ using multilingual pretrained models \citep{devlin-etal-2019-bert,conneau-etal-2020-unsupervised}, and calculates the similarity matrix $S$, where $S_{i,j}$ represents the cosine similarity between tokens $s_i$ and $t_j$. 
The \texttt{argmax} inference algorithm selects position pairs $\langle i, j\rangle$, where $S_{i,j}$ is both horizontal and vertical maximum, and outputs the word pairs corresponding to such position pairs as the word alignment. 
In this work, we use XLM-R \cite{conneau-etal-2020-unsupervised} based SimAlign with the \texttt{argmax} algorithm to extract word alignment for \ddp. 

It is worth noting that pretrained multilingual models usually use byte-pair encodings \citep[BPEs;][]{gage1994new}, a more fine-grained level than words, for tokenization. 
The \texttt{argmax} algorithm may therefore generate many-to-one alignments.
More details can be found in \citet{jalili-sabet-etal-2020-simalign}. 

Unlike bitext based word alignment methods such as GIZA++ \citep{och-ney-2003-systematic} and \texttt{fast\_align} \citep{dyer-etal-2013-simple}, SimAlign does not require any bitext pairs
to produce high quality alignments, and is therefore more suitable for the low-resource scenario with very few bitext pairs available. 

\subsection{Preliminaries}
\label{sec:preliminary}

\paragraph{Dependency annotations in $L_1$.} 
We adapt the most common data settings for supervised dependency parsing, where sentences with dependency annotations are provided. 
For a sentence $\langle w_1, \ldots, w_n \rangle$, there is one root word; every other word $w_i$ is labeled with $h_i$ and $r_i$, denoting that the head of $w_i$ is $w_{h_i}$, with the dependency relation $r_i$. 
We use these annotations to train an $L_1$ bi-affine dependency parser $\mathcal{P}_1$, following the procedure described in \S\ref{sec:background}. 

\paragraph{Bitext.} We denote the available $m$ pairs of bitext by $\mathcal{B} = \{\langle s^{(k)}, t^{(k)}\rangle\}_{k=1}^m$, where $\{s^{(k)}\}$ and $\{t^{(k)}\}$ are sentences in $L_1$ and $L_2$ respectively. 

\paragraph{Word alignment.} 
For a pair of bitext $\langle s, t\rangle$, we generate the word alignment matrix $\bm\tilde{A} \in \{0, 1\}^{|s|\times|t|}$ with SimAlign,
where $\bm{\tilde{A}}_{i,j}=1$ denotes that there exists an alignment between $s_i$ and $t_j$.

We would like the word alignment matrices to be right stochastic, which satisfy (1) that each element is non-negative and (2) that each row sums to one, to ensure that the results after projection remain distributions. 
There may exist some words that have zero or more than one aligned words in the other language; therefore, we introduce the following two matrix operators.  

\textbf{The \textit{add-dummy-column} operator $\Delta(\cdot)$}:
\begin{align*}    
    \Delta&: \mathbb{R}^{r\times c} \rightarrow \mathbb{R}^{r\times(c+1)} (\forall r, c \in \mathbb{N}_+) \\
    \Delta(\bm{M})_{i,j} &= \left\{\begin{aligned} & \bm{M}_{i,j} && (j \leq c) \\
    & 1 && (j=c+1, \sum_{k=1}^c \bm{M}_{i,k} = 0) \\
    & 0 && (j=c+1, \sum_{k=1}^c \bm{M}_{i,k} \neq 0). 
    \end{aligned}
    \right.
\end{align*}

\textbf{The \textit{row normalization} operator $\mathcal{N}^{\mathcal{R}}(\cdot)$}: 
\begin{align*}
    \mathcal{N}^{\mathcal{R}}:& \mathbb{R}^{r\times c} \rightarrow \mathbb{R}^{r\times c} (\forall r, c \in \mathbb{N}_+) \\
    \mathcal{N}^{\mathcal{R}}(\bm{M})_{i, j} &= \frac{\bm{M}_{i, j}}{\sum_{\ell} \bm{M}_{i, \ell}}. 
\end{align*}

Intuitively, the added dummy column corresponds to a \emph{null} word in word alignment literature \interalia{dyer-etal-2013-simple,schulz-etal-2016-word,jalili-sabet-etal-2020-simalign}.
We denote the source-to-target alignment matrix by $\bm{A}^{s\rightarrow t} = \mathcal{N}^\mathcal{R}\left(\Delta(\tilde{\bm A})\right)$, and the target-to-source alignment matrix by $\bm{A}^{t\rightarrow s} = \mathcal{N}^\mathcal{R}\left(\Delta(\tilde{\bm A}^\intercal)\right)$. 
Both are right stochastic matrices by definition.

\subsection{Dependency Distribution Projection}
\label{sec:ddp}
\paragraph{Arc distribution projection. } We consider a pair of bitext $\langle s, t\rangle$.
Let $P_1(s_j \mid s_i)$ denote the arc probability produced by the parser $\mathcal{P}_1$.
\begin{align*}
    P_1(s_i \mid s_{|s|+1}) &= 0. 
\end{align*}
We project $P_1(\cdot \mid \cdot)$ to $\hat{P}_2(t_q\mid t_p)$, the arc probability distributions in the parallel $L_2$ example $t$, by 
\begin{align}
    \hat{P}_2(t_q\mid t_p) = \sum_{i=1}^{|s|+1} \sum_{j=1}^{|s|} \bm{A}^{t\rightarrow s}_{p,i} P_1(s_j\mid s_i) \bm{A}^{s\rightarrow t}_{j,q}. \label{eq:3-matrices-prob}
\end{align}
It is guaranteed that $\hat{P}_2(\cdot \mid t_p)$ is a distribution for any $t_p$ -- a proof can be found in Appendix~\ref{sec:proof-arc-distribution}. 
Note that if we adopt matrix notations, where we denote $\hat{P}_2(t_q\mid t_p)$ by $\bm{\hat{P}}^{(2)}_{p, q}$ and denote $P_1(s_j\mid s_i)$ by $\bm{P}^{(1)}_{i, j}$, Eq~\eqref{eq:3-matrices-prob} is equivalent to
\begin{align*}
    \bm{\hat{P}}^{(2)} = \bm{A}^{t\rightarrow s}\bm{P}^{(1)}\bm{A}^{s\rightarrow t}. 
\end{align*}

\paragraph{Label distribution projection. } Let $P_1(\ell \mid s_j\rightarrow s_i)$ denote the label probability produced by $\mathcal{P}_1$. 
For dummy positions, we simply add a uniform distribution, that is, 
\begin{align*}
    P_1(\ell \mid s_j \rightarrow s_i) = \frac{1}{L} && \text{if $i$ or $j=|s|+1$.}
\end{align*}
We project $P_1(\cdot \mid \cdot \rightarrow \cdot)$ to $\hat{P}_2(\ell\mid t_q\rightarrow t_p)$, the label probability distributions in the parallel $L_2$ example $t$, by 
\begin{align*}
    & \hat{P}_2(\ell \mid t_q\rightarrow t_p) \\
    =&\sum_{i=1}^{|s|+1}\sum_{j=1}^{|s|+1} \bm{A}^{t\rightarrow s}_{p,i}P_1(\ell\mid s_j\rightarrow s_i)\bm{A}^{t\rightarrow s}_{q,j}
\end{align*}
It can be proved that $\hat{P}_2(\cdot \mid t_q\rightarrow t_p)$ is a distribution for any pair of $t_p$ and $t_q$. 
A detailed proof can be found in Appendix~\ref{sec:proof-label-distribution}. 

\subsection{Optimization}
\label{sec:optimization}
We train another bi-affine dependency parser $\mathcal{P}_2$ on language $L_2$, by minimizing the cross entropy between its produced probability $P_2$ against the soft silver labels $\hat{P}_2$. 
Note that the added dummy word denoting the null alignment is not eventually used in the final dependency inference process and may introduce extra noise to the model, so we instead calculate \textit{partial} cross entropy loss, which does not consider elements involving dummy words. 
Concretely, we compute the partial arc cross entropy loss for one example $t$ as follows:
\begin{align*}
    &\mathcal{L}_\textit{arc}^{(t)}(P_2, \hat{P}_2) \\
    = &-\sum_{p=1}^{|t|}\sum_{q=1}^{|t|}\hat{P}_2(t_q \mid t_p)\log P_2(t_q \mid t_p)
\end{align*}
Similarly, the partial label cross entropy loss can be computed as follows:
\begin{align*}
    \mathcal{L}_\textit{label}^{(t)}(P_2, \hat{P}_2) 
    & = -\sum_{\ell=1}^{|L|}\sum_{p=1}^{|t|}\sum_{q=1}^{|t|}\\
    & \hat{P}_2(\ell\mid t_q\rightarrow t_p)\log P_2(\ell \mid t_q\rightarrow t_p)
\end{align*}
The final loss is defined by the sum of $\mathcal{L}_\textit{arc}$ and $\mathcal{L}_\textit{label}$, that is, we train the parameters of $\mathcal{P}_2$ to minimize 
\begin{align}
    \sum_{\langle s,t \rangle \in \mathcal{B}} \mathcal{L}_\textit{arc}^{(t)}(P_2, \hat{P}_2) + \mathcal{L}_\textit{label}^{(t)}(P_2, \hat{P}_2). \label{eq:objective} 
\end{align}
More explanation about the intuition of \ddp can be found in Appendices~\ref{sec:properties} and \ref{sec:intuition}.  

\section{Experiments}
\newcommand{\emptyresult}{---}
\begin{table*}[t!]
    \centering \small \addtolength{\tabcolsep}{-4pt}
    \begin{tabular}{lccccccccccccccccc|ccccccccccccccccc}
        \toprule
        && \multicolumn{15}{c}{\bf LAS} && \multicolumn{16}{c}{\bf UAS}\\ 
        \cmidrule{3-17} \cmidrule{20-34}
        \bf Method && \multicolumn{7}{c}{\it distant languages} && \multicolumn{7}{c}{\it nearby languages} &&& \multicolumn{7}{c}{\it distant languages} && \multicolumn{7}{c}{\it nearby languages} \\
        \cmidrule{3-9} \cmidrule{11-17} \cmidrule{20-26} \cmidrule{28-34}
        && ar && hi && ko && tr && de && es && fr && it &&& ar && hi && ko && tr && de && es && fr && it  \\
        \midrule
        \citeauthor{meng-etal-2019-target} && \emptyresult && \emptyresult && \emptyresult && \emptyresult && \emptyresult && \emptyresult && \emptyresult && \emptyresult &&& 47.3 && 52.4 && 37.1 && 35.2 && 70.8 && 75.8 && 79.1 && 82.0\\
        \citeauthor{he-etal-2019-cross}   && \emptyresult && \emptyresult && \emptyresult && \emptyresult && \emptyresult && \emptyresult && \emptyresult && \emptyresult &&& 55.4 && 33.2 && 37.0 && 36.1 && 69.5 && 64.3 && 67.7 && 70.7 \\
        \citeauthor{ahmad-etal-2019-cross} && 27.9 && 28.0 && 16.1 && \emptyresult && 61.8 && 65.8 && 73.3 && 75.6 &&& 27.9 && 28.0 && 16.1 && \emptyresult && 61.8 && 65.8 && 73.3 && 75.6 \\
        \citeauthor{kurniawan-etal-2021-ppt} && 38.5 && 28.3 && 16.1 && 20.6 && 63.5 && 69.2 && \textbf{74.5} && \textbf{77.7} &&& 48.3 && 36.4 && 34.6 && 38.4 && 74.1 && 78.3 && 80.6 && 83.7 \\
        \ddp (ours) && \textbf{41.3} && \textbf{38.9} && \textbf{31.2} && \textbf{33.5} && \textbf{71.7} && \textbf{70.4} && 71.0 && 75.0 &&& \textbf{63.8} && \textbf{58.3} && \textbf{54.3} && \textbf{56.9} && \textbf{82.8} && \textbf{83.9} && \textbf{84.8} && \textbf{88.2}\\
        \bottomrule
    \end{tabular}
    \caption{
    Labeled attachment scores (LAS) and unlabeled attachment scores (UAS) on the Universal Dependencies v2.2 \citep{nivre-etal-2020-universal} standard test set, transferring from English. Following the protocol proposed by \citet{kurniawan-etal-2021-ppt}, our results are averaged across 5 runs with different random seeds; 
    The best number in each column is in boldface. 
    }
    \label{tab:ud2.2-fully-unsup}
\end{table*}

\subsection{Setup}
Throughout all experiments, the subword representation is a weighted sum of layer-size representation from a frozen pretrained model, where each layer associates with a scalar weight optimized together with other network parameters to minimize Eq~\eqref{eq:objective}. 
We convert subword features to word features by endpoint concatenation, following \citet{toshniwal-etal-2020-cross}.
We use an Adam optimizer \citep{kingma2015adam} to train all the models, where the source language parser is trained for 100 epochs with an initial learning rate $2\times 10^{-3}$ following the baseline implementation by \citet{zhang-etal-2020-efficient},\footnote{\url{https://github.com/yzhangcs/parser/tree/d7b6ae5498bd045e34c1e1c55ab8e619cf4ad353}. By grid search on learning rate and dropout ratio for source parser training, we verify that their default hyperparameters are always among the best performing ones on the Universal Dependencies v2.2 English development set. Therefore, we adopt their hyperparameter choices on learning rate and dropout ratio. } and the target language parser is trained for 30 epochs with an initial learning rate $5\times 10^{-4}$.\footnote{We do not observe further training loss decrease when training for more epochs. The learning rate for \ddp is tuned to optimize the development loss for German. }
We use the loss against silver projected distributions on the development set for \ddp and the development LAS against projected trees for baselines for early stopping.\footnote{\ddp does not provide a set of discrete silver trees for LAS and UAS calculation. } 
For evaluation, we ignore all punctuation following the most common convention \interalia{ma-xia-2014-unsupervised,rasooli-collins-2015-density,kurniawan-etal-2021-ppt}.

If not specified, 
\begin{itemize}[leftmargin=*,itemsep=-1mm]
    \item All models in target languages are initialized with the trained source language parser. 
    \item All word alignments are obtained by XLM-R based SimAlign \citep{jalili-sabet-etal-2020-simalign}, using BPE tokenization and the \texttt{argmax} algorithm.
    \item XLM-R is used as the feature extractor. 
\end{itemize}
For analysis purposes, we report the performance on the standard development sets to avoid tuning on the test sets.
\subsection{Results: Fully Unsupervised Transfer}

We compare \ddp to prior work in the minimal annotation settings (Table~\ref{tab:ud2.2-fully-unsup}), where an English dependency treebank is the only annotation that involves human effort.
We select target languages from the overlap between those considered by \citet{kurniawan-etal-2021-ppt}, those covered by XLM-R \citep{conneau-etal-2020-unsupervised} training corpora, and those supported by CRISS \citep{tran-etal-2020-cross}, resulting in eight languages: Arabic (ar), Hindi (hi), Korean (ko), Turkish (tr), German (de), Spanish (es), French (fr), and Italian (it).

We generate translations of English treebank sentences using CRISS to construct the required bitext. 
To ensure the quality of the unsupervised bitext, we discard (1) translations where at least 80\% of words appear in the corresponding source sentences, which are likely to be copies, (2) those containing a CRISS language token other than the target language, which are likely to be false translation into another language and (3) those with 80\% or more words appearing in the translated sentence more than once, which are likely to be repetitions. 

Transferring from an English parser, \ddp achieves the best UAS across all eight target languages, 
and the best LAS on six languages out of eight. 
In addition, we find that \ddp is consistent across random seeds, with a standard deviation less than $0.8$ for every corresponding number in Table~\ref{tab:ud2.2-fully-unsup}. 

\subsection{Ablation Study}
\begin{figure*}[t]
\includegraphics[width=0.27\textwidth]{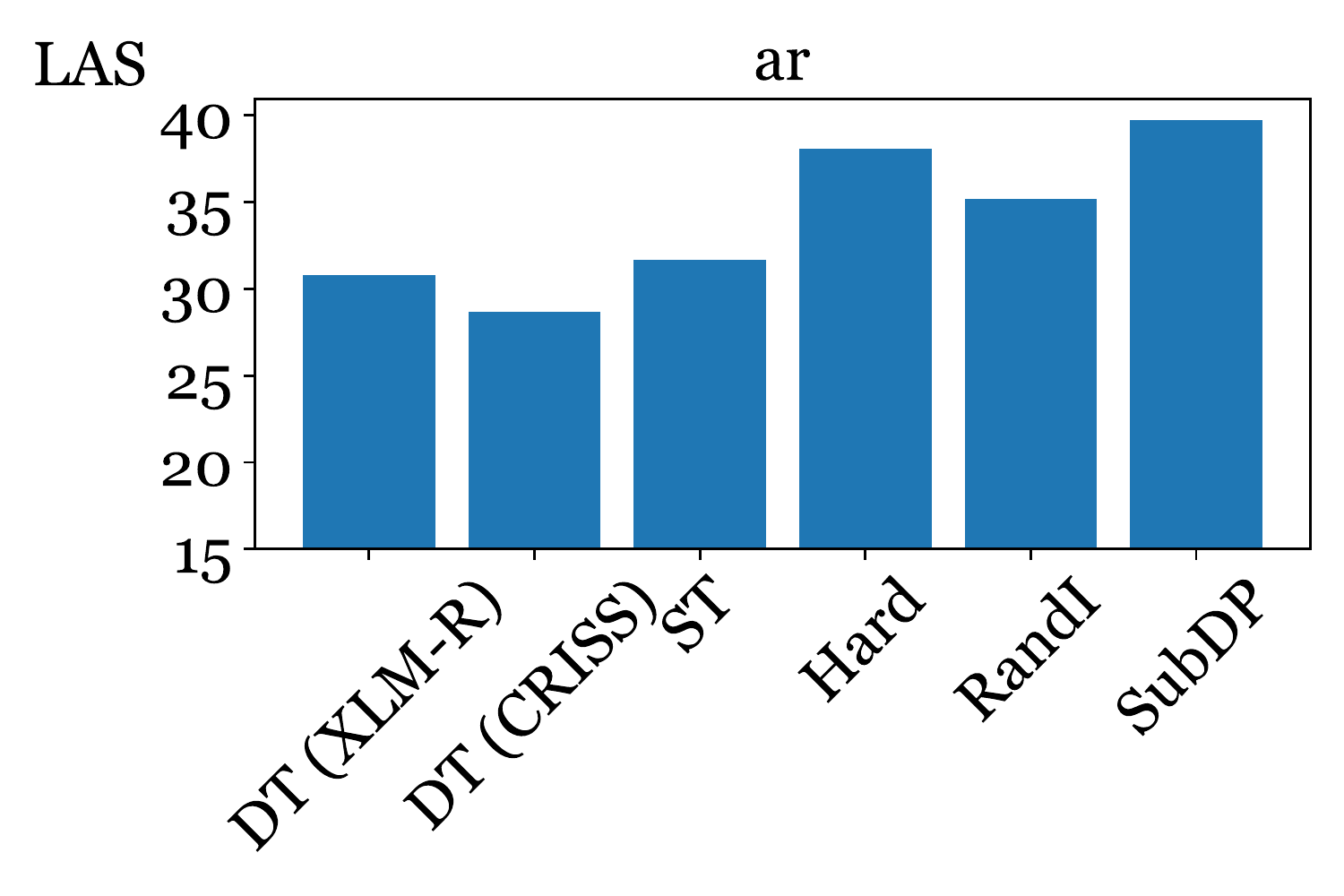} \hspace{-25pt}
\includegraphics[width=0.27\textwidth]{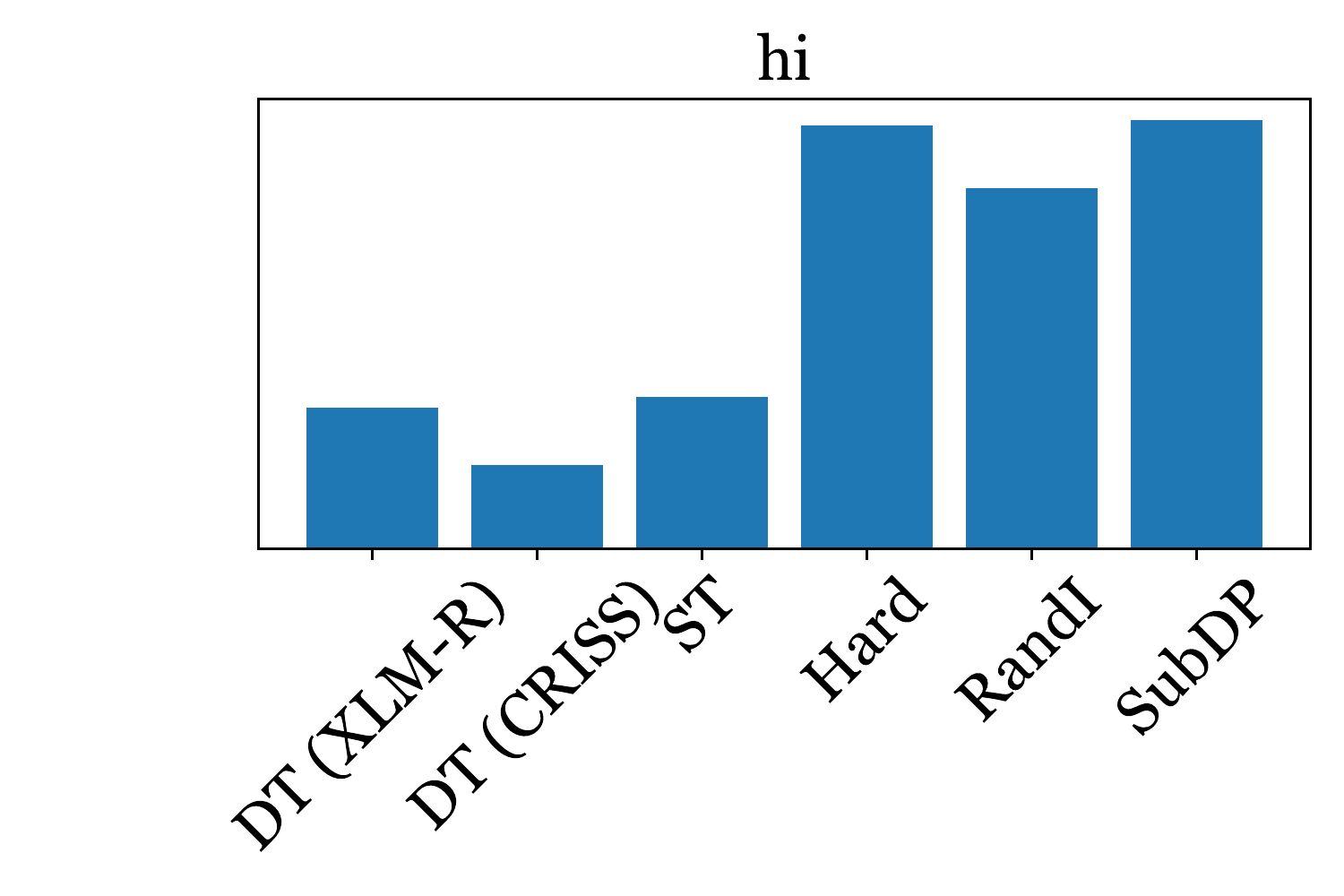} \hspace{-25pt}
\includegraphics[width=0.27\textwidth]{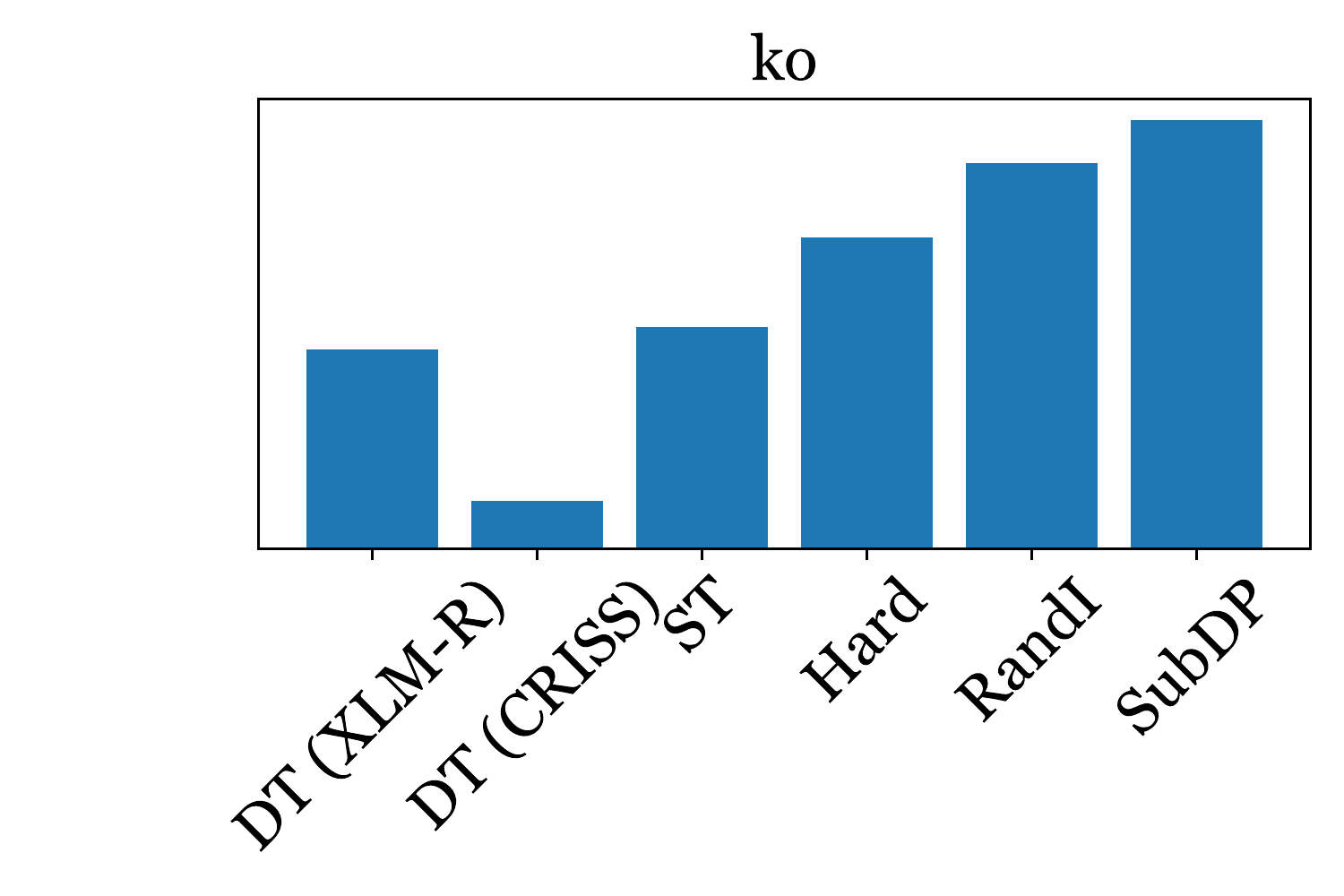} \hspace{-25pt}
\includegraphics[width=0.27\textwidth]{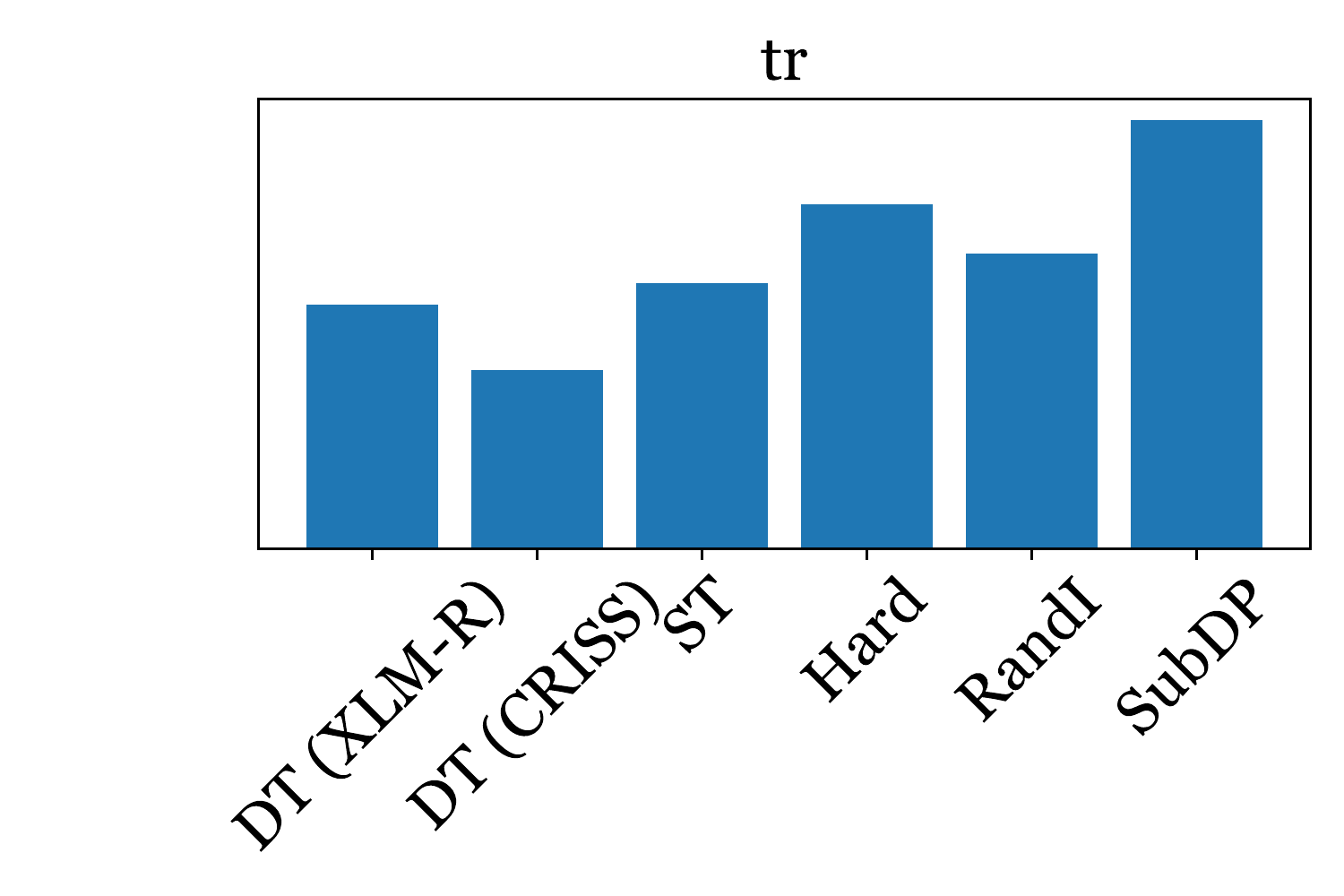} \\
\includegraphics[width=0.27\textwidth]{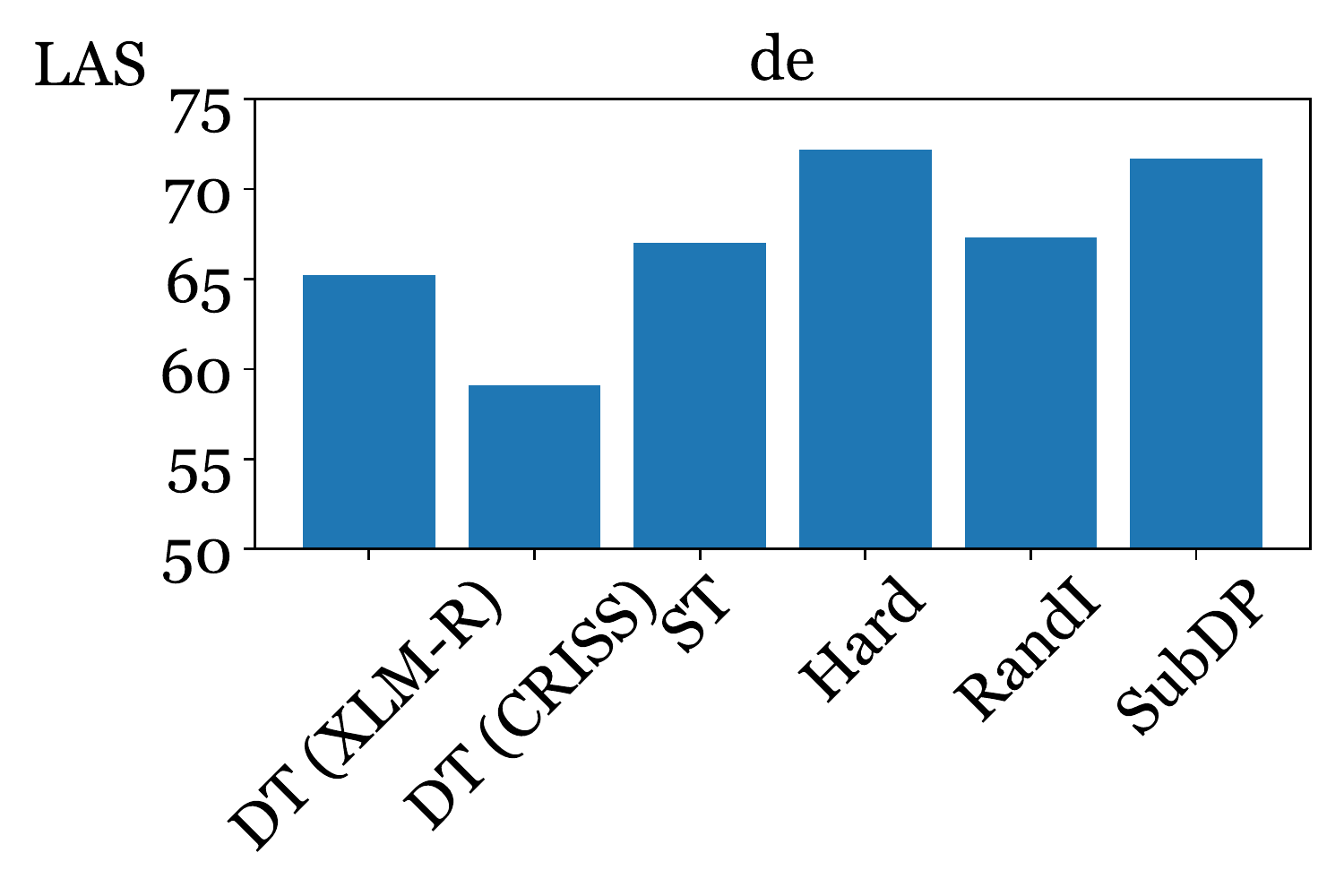} \hspace{-25pt}
\includegraphics[width=0.27\textwidth]{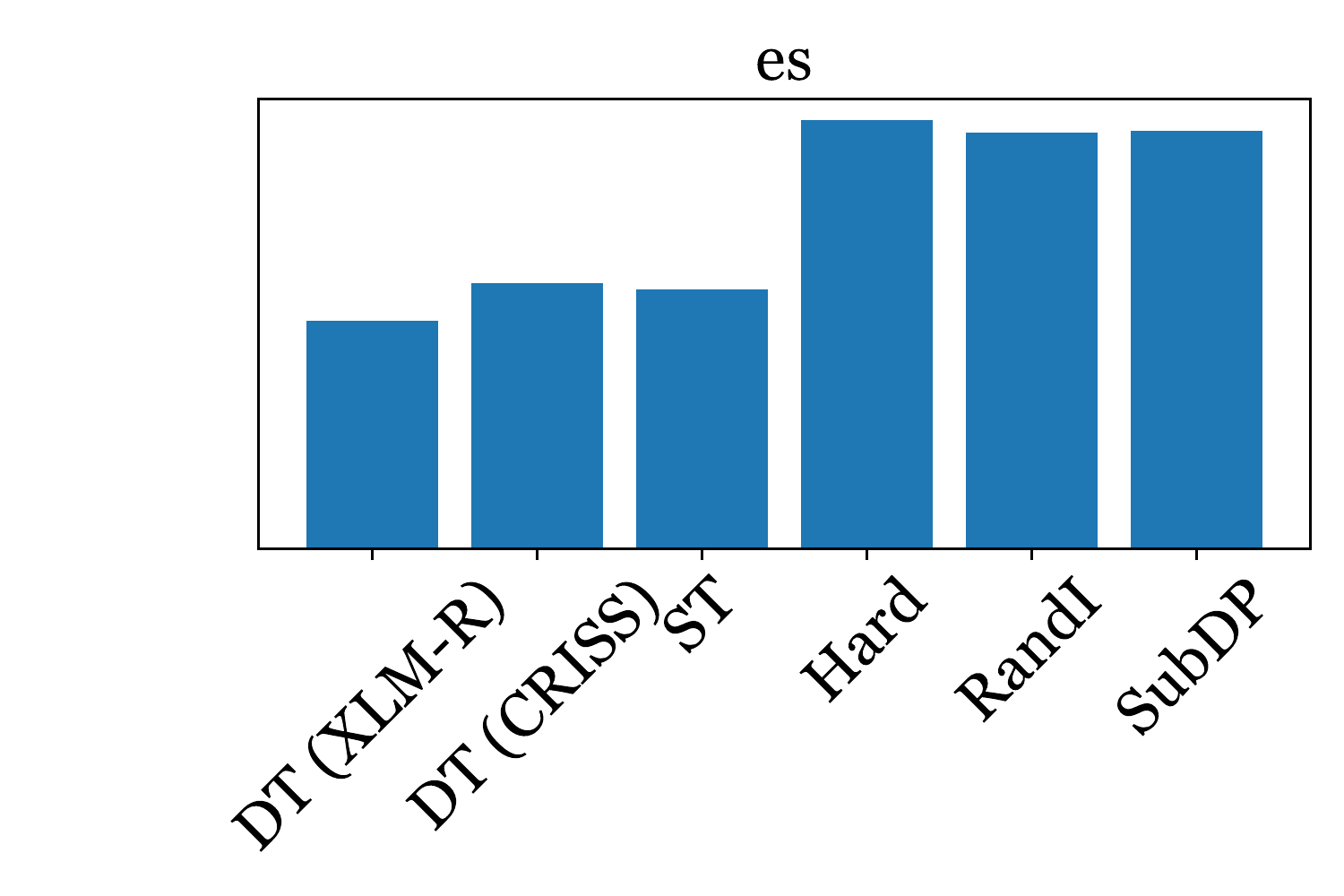} \hspace{-25pt}
\includegraphics[width=0.27\textwidth]{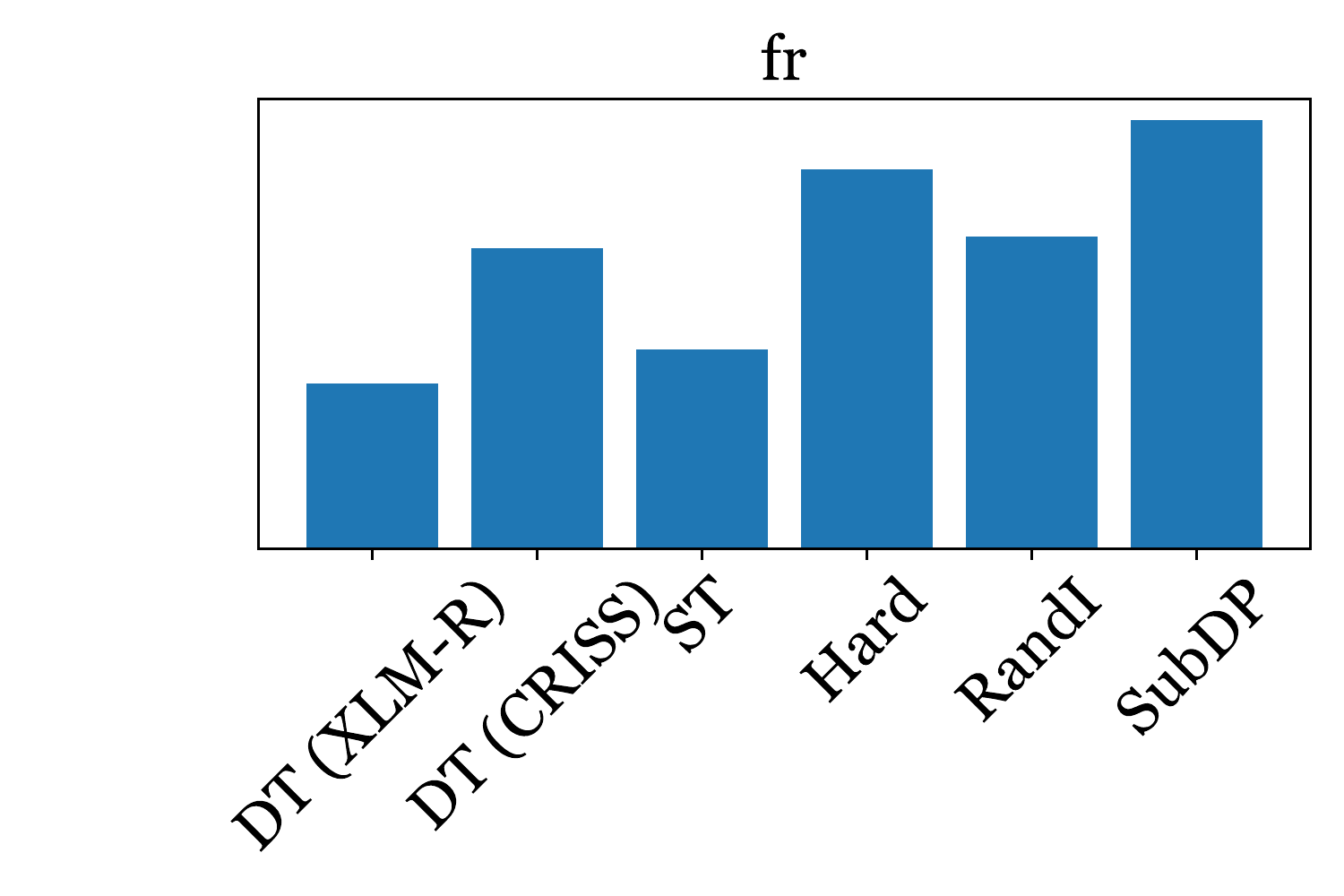} \hspace{-25pt}
\includegraphics[width=0.27\textwidth]{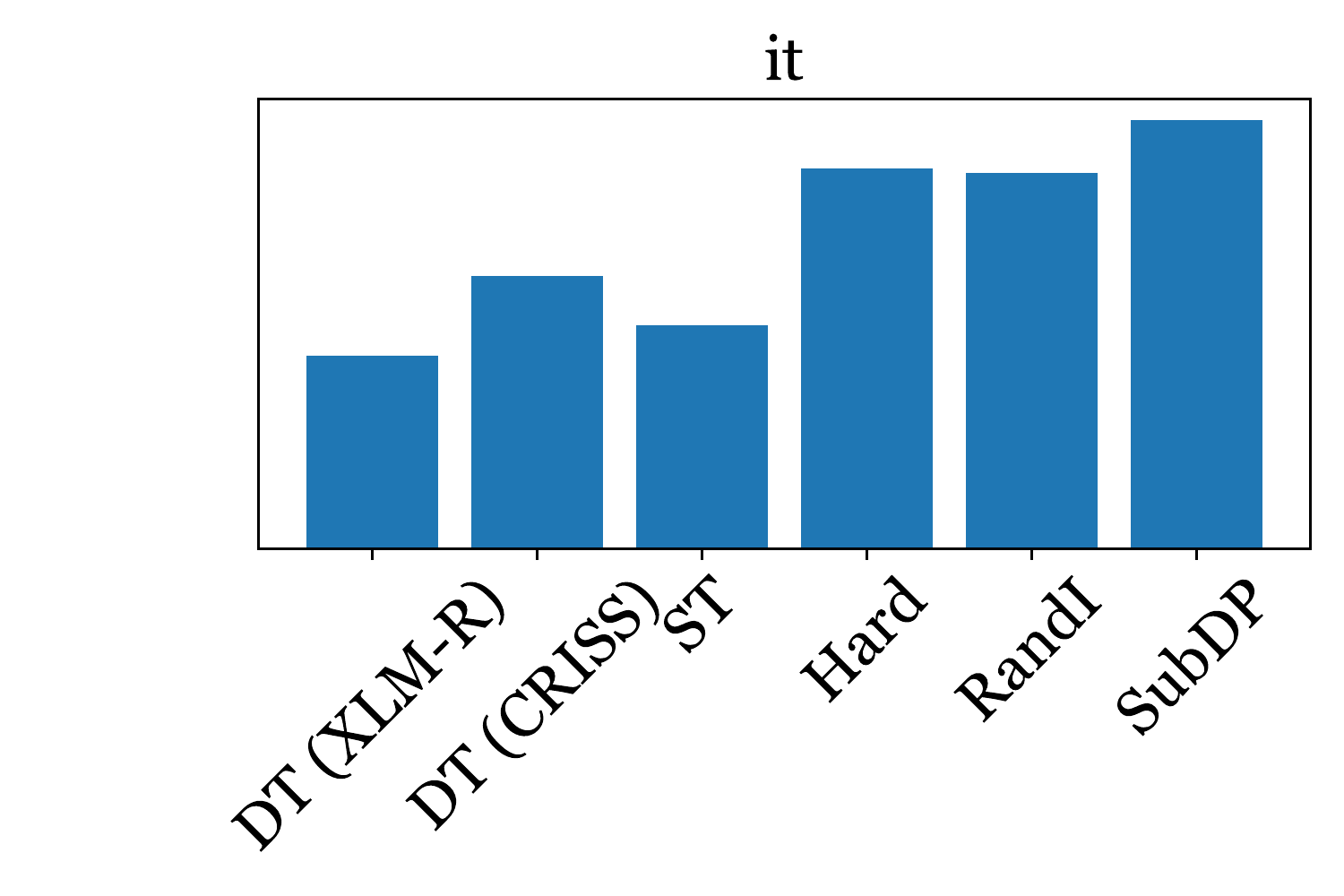} 
\caption{\label{fig:ablation} LAS on the Universal Dependencies v2.2 standard development set. All numbers are averaged across 5 runs. Corresponding UAS plots can be found in Appendix~\ref{sec:ablation-uas}.}
\end{figure*}
We introduce the following baselines with the same annotated data availability for ablation study: 

\begin{enumerate}[leftmargin=*,itemsep=0mm]
\item \textbf{Direct transfer of English models (DT).} 
We train a bi-affine dependency parser on annotated English dependency parsing data, and directly test the model on other languages. 
This approach is expected to outperform a random baseline as it has a pretrained cross-lingual language model-based feature extractor, which may implicitly enable cross-lingual transfer. 
For this baseline, we test both XLM-R and CRISS encoder as the feature extractor, as \ddp benefits from both models. 
\item \textbf{Self-training (ST). } Following \citet{kurniawan-etal-2021-ppt}, we consider self-training as another baseline. 
Starting by applying a DT parser to the target language, we train another parser fit the predicted trees. 
\item \textbf{Hard projection of treebank (Hard).} It is intuitive to compare \ddp against the hard tree projection baseline \citep{lacroix-etal-2016-frustratingly}, where we use the same set of bitext and alignments to project trees to the target languages, keeping only confident edges (i.e., edges with both sides aligned in a one-to-one alignment).
We use the projected trees to train a parser in the target language. 
It is worth noting that this baseline falls into the category of treebank projection (\S\ref{sec:related}), instead of annotation projection, as we directly translate English treebank sentences with CRISS.
\item \textbf{Random target parser initialization (RandI). } Instead of using the trained English model as the initialization of target parsers, we randomly initialize the weights in this baseline. 
\end{enumerate} 
All of the baselines use bi-affine dependency parsers, with pretrained cross-lingual language models (XLM-R or CRISS) as feature extractors. \\[0pt]

We compare the LAS between \ddp and the baselines above (Figure~\ref{fig:ablation}), and find that 
\begin{itemize} [leftmargin=*,itemsep=0mm]
\item Across all languages, \ddp significantly outperforms DT with either XLM-R or CRISS word feature extractor. 
ST does improves over DT consistently, but is much less competitive than \ddp. 
This indicates that the gain of \ddp over prior work is not simply from more powerful word features.
\item While hard treebank projection using the method proposed by \citet{lacroix-etal-2016-frustratingly} is quite competitive, \ddp consistently produce competitive (Hindi, German, Spanish) or better (Arabic, Korean, Turkish, French, Italian) results. 
\item Initializing the target language parser with a trained source language (English in our experiments) parser is helpful to improve performance across the board, and therefore should be considered as a general step in future work on zero-shot cross-lingual dependency parsing. 
\end{itemize}

\subsection{Analysis: Effect of Alignment Methods}
\begin{table}[t]
    \centering
    \small
    \begin{tabular}{lcc|cc}
        \toprule
        \bf Lang. & \multicolumn{2}{c|}{\tt BPE argmax} & \multicolumn{2}{c}{\tt 1:1 only}\\
        & LAS & UAS & LAS & UAS \\
        \midrule
        ar &  39.7 & 60.7 & \bf 40.2 & \bf 61.1 \\
        hi & \bf 39.7 & \bf 57.4 & 38.7 & 56.5 \\
        ko & \bf 31.1 & \bf 51.3 & 27.3 & 49.6\\
        tr & \bf 37.8 & \bf 56.7 & 33.3 & 55.8 \\
        \midrule
        avg. dist. & \bf 37.1 & \bf 56.5 & 34.8 & 55.8 \\
        \midrule
        \midrule 
        de & 71.7 & 81.6 & \bf 72.6 & \bf 83.8 \\
        es & 67.3 & 79.7 & \bf 70.4 & \bf 84.2 \\
        fr & 71.8 & 85.3 & \bf 72.6 & \bf 87.7 \\ 
        it & 74.6 & 85.9 & \bf 76.0 & \bf 88.8 \\
        \midrule 
        avg. nearby & 71.4 & 83.1 & \bf 72.9 & \bf 86.1 \\
        \bottomrule
    \end{tabular}
    \caption{LAS and UAS on the Universal Dependencies v2.2 \citep{nivre-etal-2020-universal} standard development set, averaged across 5 runs with different random seeds. 1:1 only denotes the filtered one-to-one alignments. The best LAS and UAS for each language is bolded. }
    \label{tab:alignments}
\end{table}
Since most existing work has used only one-to-one alignment for for annotation projection 
\interalia{ma-xia-2014-unsupervised,lacroix-etal-2016-frustratingly,rasooli-etal-2021-wikily}, we would like to analyze the effect of introducing many-to-one alignment edges in \ddp. 
We filter SimAlign BPE \texttt{argmax} to obtain a more conservative version, dropping all many-to-one edges (i.e., those have a word linked to multiple edges),\footnote{~This approach is different from Hard as it projects the source tree distribution, instead of a concrete tree, to the target language, yielding soft target trees as silver labels to train target language parsers.} 
and compare it to the BPE \texttt{argmax} algorithm (Table~\ref{tab:alignments}).

While the confident one-to-one alignment achieves further improvement on Arabic and all four nearby languages, we find that the many-to-one BPE \texttt{argmax} alignment is important to the superior transfer performance on Hindi, Korean, and Turkish. 
Given the fact that the scores are quite similar for Arabic, the results generally suggest using the many-to-one SimAlign BPE \texttt{argmax} alignments for transferring from English to distant languages, while using the more confident one-to-one alignments for nearby languages.

\subsection{Results: Multiple Source Languages}
\begin{table}[t]
    \centering \small 
    \begin{tabular}{lcccc}
        \toprule 
         \bf Method & de & es & fr & it \\
         \midrule
         \citet{zhang-barzilay-2015-hierarchical} & 62.5 & 78.0 & \bf 78.9 & 79.3 \\
         \citet{guo2016representation} & 65.0 & 79.0 & 77.7 & 78.5 \\
         \citet{schuster-etal-2019-cross}$^\ddagger$ & 61.7 & 76.6 & 76.3 & 77.1\\
         DT (XLM-R)$^{\ddagger,*}$ & 73.1 & 82.2 & 75.5 & 79.5 \\
         \ddp (XLM-R)$^{\ddagger,*}$ & \bf 78.5 & 72.1 & 73.1 & 74.3 \\
         DT w/ \ddp init.$^{\ddagger,*}$ & 76.1 & \bf 82.6 & 77.7 & \bf 81.9\\
         \bottomrule
    \end{tabular}
    \caption{LAS on Universal Dependencies v2.0 \citep{mcdonald-etal-2013-universal} standard test set. $\ddagger$: methods with minimal annotation. $*$: results from our experiments; other results are taken from \citet{schuster-etal-2019-cross}. The best number for each language is in boldface. }
    \label{tab:multiple-source}
\end{table}
Following \citet{schuster-etal-2019-cross}, we use Universal Dependencies v2.0 \citep{mcdonald-etal-2013-universal} to evaluate zero-shot cross-lingual transfer from multiple source languages (Table~\ref{tab:multiple-source}).\footnote{We do not report performances for Portuguese and Swedish as they are not covered by CRISS; 
however, the annotated treebanks in these languages are still used for transferring to other languages.}
For each language among German (de), Spanish (es), French (fr), Italian (it), Portuguese (pt) and Swedish (sv), annotated treebanks from all other languages and English can be used for training and development purposes. 
For \ddp, we generate bitext from all applicable source languages with CRISS. 

\ddp outperforms the previous state-of-the-art on German by 13.5 LAS, but under-performs the DT baseline on the other three languages. 
However, if we start with a trained \ddp parser for a target language, and use the standard training data (i.e., treebanks in other languages) to further train a bi-affine dependency parser (DT w/ \ddp init.), 
we are able to achieve better results than DT across the board, obtaining competitive or even better LAS than methods that use extra annotations other than source treebanks \citep{zhang-barzilay-2015-hierarchical,guo2016representation}.

\subsection{Results: Transfer with Supervised Bitext}
\begin{figure*}[t!]
\centering
\includegraphics[width=0.32\textwidth]{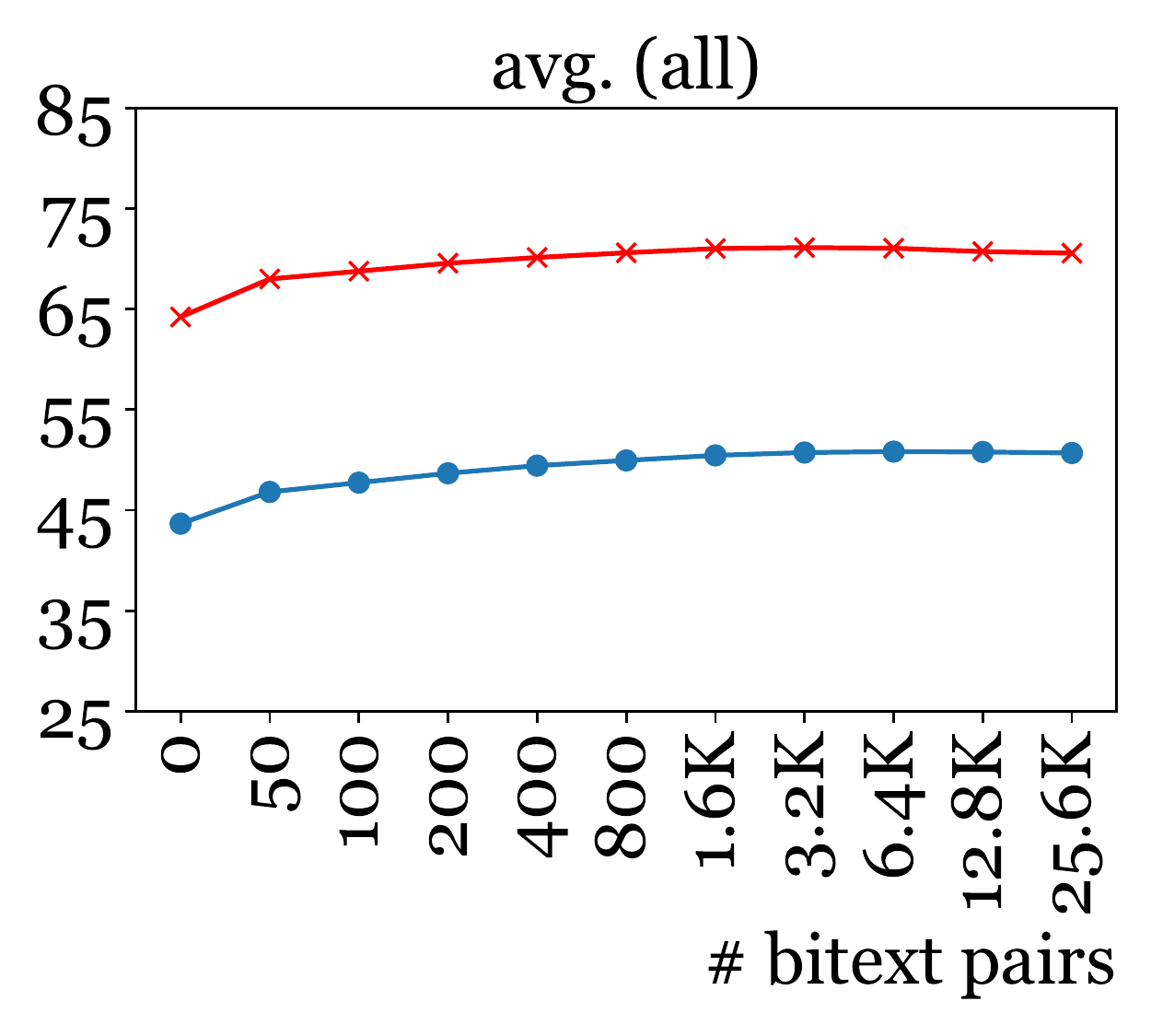}
\includegraphics[width=0.32\textwidth]{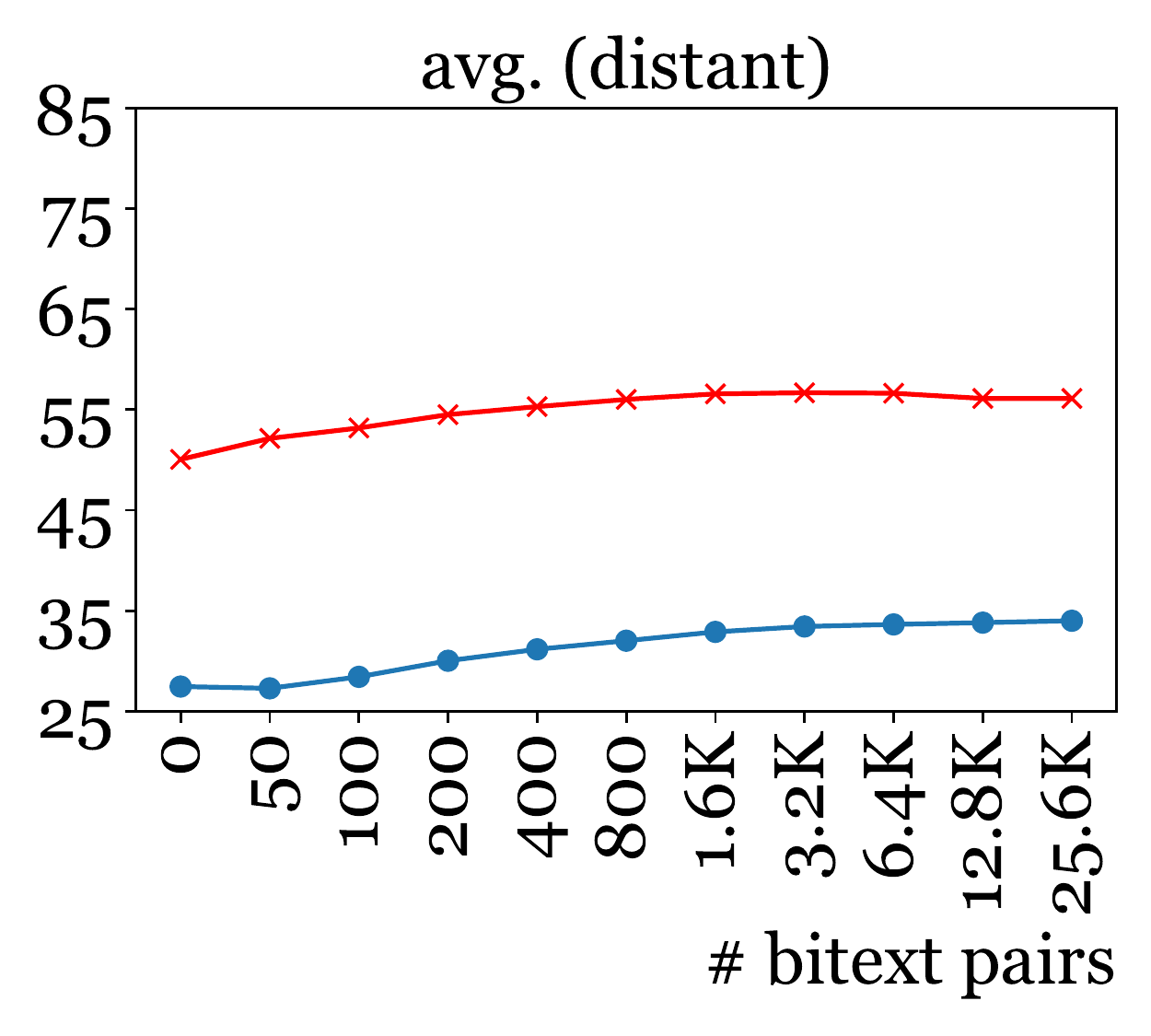}
\includegraphics[width=0.32\textwidth]{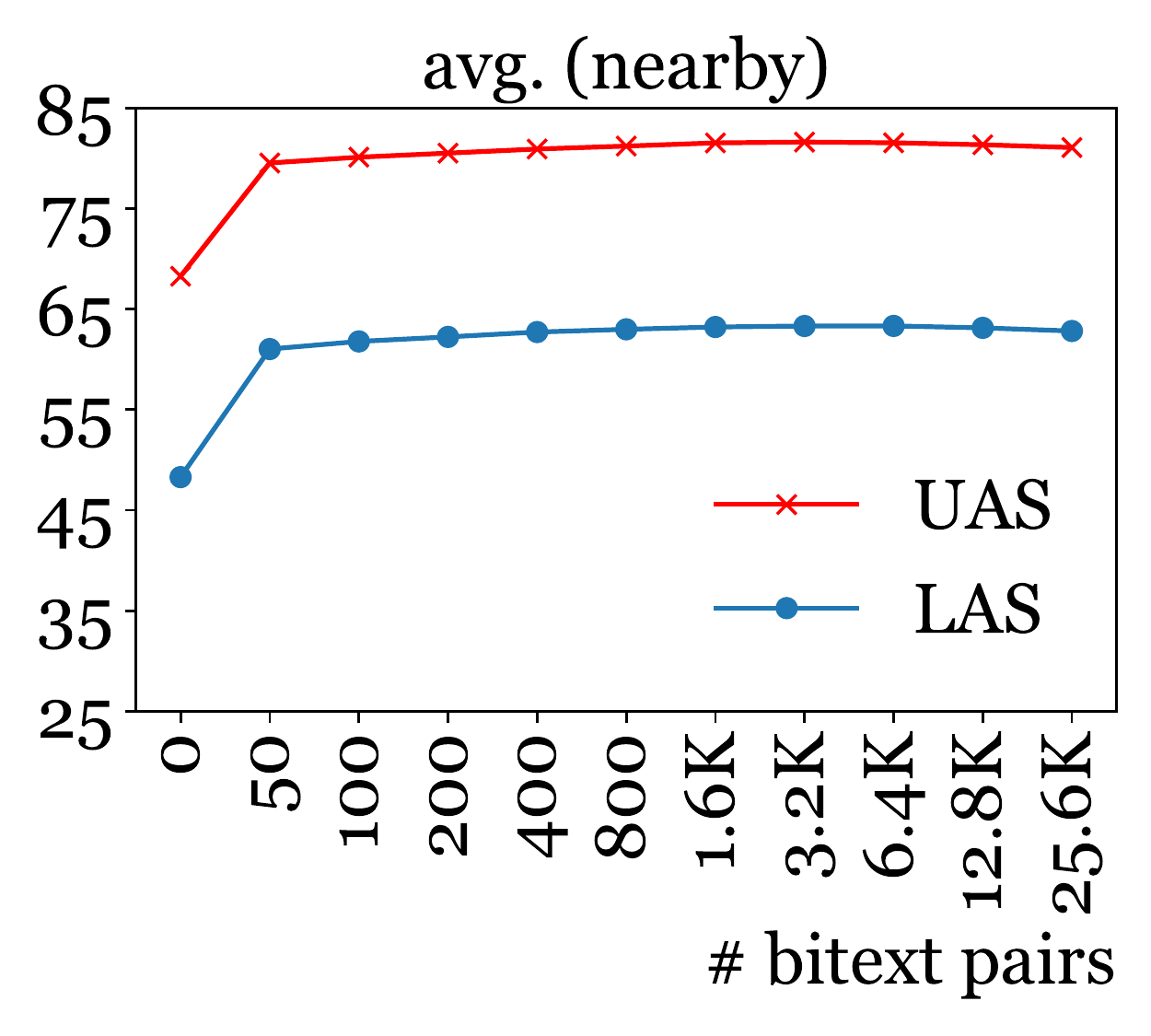}
\caption{Averaged LAS and UAS on the Universal Dependencies v2.2 standard development set, with respect to the number of bitext pairs (best viewed in color). For each language, we run 5 times with different random seeds. The $x$-axis is in log scale. All European languages are categorized as nearby languages, while the remaining are treated as distant languages. Plots for individual languages can be found in Appendix~\ref{sec:invididual-plots}. \label{fig:data-efficiency}}
\end{figure*}
We further extend \ddp to a larger set of eighteen languages, including Arabic (ar), Czech (cs), German (de), Spanish (es), Finnish (fi), French (fr), Hindi (hi), Hugarian (hu), Italian (it), Japanese (ja), Korean (ko), Norwegian (no), Portuguese (pt), Russian (ru), Tamil (ta), Telugu (te), Vietnamese (vi) and Chinese (zh). 
We transfer from English to each target language with Wikimatrix bitext \citep{schwenk-etal-2021-wikimatrix}, where the examples are mined with a encoding similarity based bitext miner trained with annotated bitext. 
We vary the number of
Wikimatrix bitext pairs, selecting the number of pairs within the geometric sequence $\{50\times 2^{k}\}_{k=0}^9$, 
leaving 10\% of the examples for development.

On average and for nearby languages (Figure~\ref{fig:data-efficiency}), we find that the performance of \ddp with 50 pairs of bitext is quite close to that with 25K pairs of bitext. 
Although some distant languages generally require more bitext for further improvement, \ddp achieves outperforms the direct transfer baseline by a nontrivial margin with a small amount (e.g., 800-1.6K pairs) of bitext. 
\section{Discussion}

In line with recent work \citep{rasooli-etal-2021-wikily} which shows that cross-lingual transfer can be done effectively with weak supervision such as Wikipedia links, we further demonstrate the potential of zero-shot cross-lingual dependency parsing with zero additional supervision, even between distant languages that do not share vocabulary or subwords.
Our work suggests a new protocol for dependency annotations of low-resource languages: (1) training a pretrained multilingual model following existing work such as XLM-R \citep{conneau-etal-2020-unsupervised} and CRISS \citep{tran-etal-2020-cross}, (2) annotate a small number of bitext pairs or generate bitext with trained unsupervised translation models, and (3) train a zero-shot cross-lingual dependency parser using \ddp. 

Our contribution to zero-shot cross-lingual dependency parsing is arguably orthogonal to contextualized representation alignment \citep{schuster-etal-2019-cross,wang-etal-2019-cross}, as we use frozen pretrained multilingual models to extract features, where they finetune these models to seek better transfer across languages. 
In addition, projection quality controls by heuristic rule--based filtering \citep{rasooli-collins-2015-density} may also be combined with \ddp to further improve the performance.

Our results, on the other hand, demonstrate that multilingual pretrained models may have more applications
beyond representation-based direct transfer---information extracted from these models without further supervision, such as word alignment in this work, may further benefit downstream tasks, such as zero-shot cross-lingual dependency parsing shown in this work, with appropriate usage. 

We suggest that \ddp can be extended to other scenarios wherever relevant parallel grounding signals are applicable, such as cross-lingual named entity recognition, cross-lingual constituency parsing or zero-shot scene graph parsing for images using only the dependency supervision in text.
We leave the further exploration of \ddp on other tasks for future work.

\bibliography{anthology,custom}
\bibliographystyle{acl_natbib}

\newpage
\appendix
\section{Proofs of the Propositions in the Main Content}
\label{sec:proofs}
In this section, we show that both $P_2(\cdot\mid\cdot)$ and $P_2(\cdot\mid \cdot\rightarrow\cdot)$ are probability distributions, where the key idea is applying the sum-product algorithm. 
\subsection{Distribution property of $P_2(\cdot\mid\cdot)$}
\label{sec:proof-arc-distribution}
\paragraph{Proposition 1} Suppose that $P_1(\cdot\mid s_i)$ is a probability distribution for any $s_i$, and that $\bm{A}^{t\rightarrow s}$ and $\bm{A}^{s\rightarrow t}$ are right-stochastic matrices (i.e., each row of the matrices defines a probability distribution). Let $P_2(t_p\mid t_q) = \sum_{i=1}^{|s|+1}\sum_{j=1}^{|s|+1} \bm{A}^{t\rightarrow s}_{p,i}P_1(s_j\mid s_i)\bm{A}^{s\rightarrow t}_{j,q}$. We have that $P_2(\cdot\mid t_p)$ is a distribution for any $t_p$. 
\paragraph{Proof.}
First, for any combination of $i, j, p, q$, we have that $\bm{A}^{t\rightarrow s}_{p, i} \geq 0$, $P_1(s_j\mid s_i) \geq 0$, $\bm{A}^{s\rightarrow t}_{j,q}\geq 0$, therefore, 
\begin{align*}
    P_2(t_q\mid t_p) = \sum_{i=1}^{|s|+1}\sum_{j=1}^{|s|+1} \bm{A}^{t\rightarrow s}_{p,i}P_1(s_j\mid s_i)\bm{A}^{s\rightarrow t}_{j,q} \geq 0
\end{align*}
On the other hand, 
\begin{align*}
    &\sum_{q=1}^{|t|+1} P_2(t_q\mid t_p) \\
    =& \sum_{q=1}^{|t|+1}\sum_{i=1}^{|s|+1}\sum_{j=1}^{|s|+1} \bm{A}^{t\rightarrow s}_{p,i}P_1(s_j\mid s_i)\bm{A}^{s\rightarrow t}_{j,q} \\
    =& \sum_{i=1}^{|s|+1}\sum_{j=1}^{|s|+1} \bm{A}^{t\rightarrow s}_{p,i} P_1(s_j\mid s_i) \left(\sum_{q=1}^{|t|+1}\bm{A}^{s\rightarrow t}_{j,q}\right) \\
    =& \sum_{i=1}^{|s|+1}\sum_{j=1}^{|s|+1} \bm{A}^{t\rightarrow s}_{p,i} P_1(s_j\mid s_i) \\
    =& \sum_{j=1}^{|s|+1} \left(\sum_{i=1}^{|s|+1}\bm{A}^{t\rightarrow s}_{p,i}\right) P_1(s_j\mid s_i)  \\
    =& \sum_{j=1}^{|s|+1}P_1(s_j\mid s_i)  \\
    =& 1.               & \square
\end{align*}

\subsection{Distribution property of $P_2(\cdot\mid \cdot\rightarrow\cdot)$}
\label{sec:proof-label-distribution}
\paragraph{Preposition 2} Suppose that $P_1(\cdot\mid s_j\rightarrow s_i)$ is a probability distribution for any combination of $s_i$ and $s_j$, and that $\bm{A}^{t\rightarrow s}$ is a right-stochastic matrix. 
Let $P_2(\ell\mid t_q \rightarrow t_p) = \sum_{i=1}^{|s|+1}\sum_{j=1}^{|s|+1}\bm{A}^{t\rightarrow s}_{p,i}P_1(\ell\mid s_j\rightarrow s_i)\bm{A}^{t\rightarrow s}_{q,j}$. 
We have that $P_2(\cdot \mid t_q \mid t_p)$ is a probability distribution for any $t_p$ and $t_q$. 

\paragraph{Proof.} Similarly to the proof in \S\ref{sec:proof-arc-distribution}, it is easy to show that for any $\ell, t_p, t_q$, 
\begin{align*}
    P_2(\ell\mid t_q\rightarrow t_p) \geq 0. 
\end{align*}

We next consider the sum over $\ell$ for a specific pair of $t_p$ and $t_q$, where we have 
\begin{align*}
    &\sum_{\ell=1}^{|L|}P_2(\ell\mid t_q\rightarrow t_p) \\
    =& \sum_{\ell=1}^{|L|}\sum_{i=1}^{|s|+1}\sum_{j=1}^{|s|+1}\bm{A}^{t\rightarrow s}_{p,i}P_1(\ell\mid s_j\rightarrow s_i)\bm{A}^{t\rightarrow s}_{q,j} \\
    =& \sum_{i=1}^{|s|+1}\sum_{j=1}^{|s|+1}\bm{A}^{t\rightarrow s}_{p,i}\bm{A}^{t\rightarrow s}_{q,j} \left( \sum_{\ell=1}^{|L|}P_1(\ell\mid s_j\rightarrow s_i)\right) \\
    =& \sum_{i=1}^{|s|+1}\sum_{j=1}^{|s|+1}\bm{A}^{t\rightarrow s}_{p,i}\bm{A}^{t\rightarrow s}_{q,j} \\
    =& \sum_{i=1}^{|s|+1}\bm{A}^{t\rightarrow s}_{p,i}\left(\sum_{j=1}^{|s|+1}\bm{A}^{t\rightarrow s}_{q,j}\right) \\
    =& \sum_{i=1}^{|s|+1}\bm{A}^{t\rightarrow s}_{p,i} \\
    =& 1.              
\end{align*}
\hfill $\square$

\section{Properties of Dependency Distribution Projection}
\label{sec:properties}
\paragraph{Preposition 3} Dependency distribution projection reduces to hard projection  \citep{lacroix-etal-2016-frustratingly} when (1) the source is a discrete parse tree, and (2) there are only one-to-one word alignment. 
\paragraph{Proof. } We prove the preposition for arc distributions here, which can be immediately generalized to label distributions due to the discreteness property. 

For a pair of bitext $\langle s, t\rangle$, under hard projection \citep{lacroix-etal-2016-frustratingly}, there exists an edge from $t_q$ to $t_p$ when and only when there exist $i, j$ such that (1) there exists an edge from $s_j$ to $s_i$, (2) $s_i$ is aligned to $t_p$, and (3) $s_j$ is aligned to $t_q$. 
It is worth noting that for any pair of $p, q$, there is at most one pair of $\langle i, j\rangle$ satisfying the above conditions (otherwise it violates the one-to-one alignment assumption).  

We consider the case of \ddp. If there exists a (unique) pair of $\langle i, j\rangle$ that satisfies all the aforementioned three conditions, we have 
\begin{align*}
    P_1(s_j\mid s_i) &= 1, \\
    \bm{A}^{t\rightarrow s}_{p, i} & = 1, && \bm{A}^{t\rightarrow s}_{p, i'} = 0 (i'\neq i),\\
    \bm{A}^{s\rightarrow t}_{j, q} & = 1, && \bm{A}^{s\rightarrow t}_{j', q} =0 (j'\neq j). 
\end{align*}
Therefore, 
\begin{align*}
    \hat{P}_2(t_q \mid t_p) &= \sum_{i''=1}^{|s|+1} \sum_{j''=1}^{|s|+1} \bm{A}^{t\rightarrow s}_{p,i''} P_1(s_{j''}\mid s_{i''}) \bm{A}^{s\rightarrow t}_{j'',q} \\ 
    &= \bm{A}^{t\rightarrow s}_{p,i} P_1(s_{j}\mid s_{i}) \bm{A}^{s\rightarrow t}_{j,q} \\ 
    &= 1
\end{align*}

On the other hand, if there do not exist a pair of $\langle i, j\rangle$ that satisfies all three conditions, for any pair of $\langle i, j\rangle$, at least one of the following is true, 
\begin{align*}
    P_1(s_j\mid s_i) &= 0, \\
    \bm{A}^{t\rightarrow s}_{p, i} & = 0,\\
    \bm{A}^{s\rightarrow t}_{j, q} & = 0.
\end{align*}
Therefore, 
\begin{align*}
    \hat{P}_2(t_q \mid t_p) &= \sum_{i''=1}^{|s|+1} \sum_{j''=1}^{|s|+1} \bm{A}^{t\rightarrow s}_{p,i''} P_1(s_{j''}\mid s_{i''}) \bm{A}^{s\rightarrow t}_{j'',q} \\ 
    &= 0. 
\end{align*}
That is, \ddp has the same behavior as \citet{lacroix-etal-2016-frustratingly} under the given assumptions.  \hfill $\square$ 

\paragraph{Preposition 4} Given a discrete source tree, \ddp assigns non-zero probability to any dependency arc generated by hard projection \citep{lacroix-etal-2016-frustratingly}. 
\paragraph{Proof. } Similarly to the proof to Preposition 3, if hard projection generates an arc $t_q\rightarrow t_p$, there exists a pair of $\langle i, j\rangle$ such that 
\begin{align*}
    P_1(s_j\mid s_i) &= 1, \\
    \bm{\tilde{A}}_{i, p} & = 1 \Rightarrow \bm{A}^{t\rightarrow s}_{p, i} > 0,\\
    \bm{\tilde{A}}_{j, q} & = 1 \Rightarrow \bm{A}^{s\rightarrow t}_{j, q} > 0,
\end{align*}
Therefore, 
\begin{align*}
    \hat{P}_2(t_q \mid t_p) &= \sum_{i''=1}^{|s|+1} \sum_{j''=1}^{|s|+1} \bm{A}^{t\rightarrow s}_{p,i''} P_1(s_{j''}\mid s_{i''}) \bm{A}^{s\rightarrow t}_{j'',q} \\ 
    &\geq  \bm{A}^{t\rightarrow s}_{p,i} P_1(s_{j}\mid s_{i}) \bm{A}^{s\rightarrow t}_{j,q} > 0. 
\end{align*}
This can be immediately generalized to label distribution due to the discreteness of the input tree. 
\hfill $\square$

\section{Intuition on Dummy Column and Partial Cross Entropy}
\label{sec:intuition}
In this section, we provide more intuition on the added dummy column (\S\ref{sec:ddp}), and the partial cross entropy optimization (\S\ref{sec:optimization}) used in \ddp. 

Consider an alternative and intuitive approach $\mathcal{A}$, which projects a source tree distribution by the following steps, taking arc distribution projection as an example: 
\begin{enumerate}[leftmargin=*,itemsep=0mm]
    \item Given $\tilde{\bm{A}}$, obtain source-to-target and target-to-source alignment matrices $\bm{\bar{A}}^{s\rightarrow t} = \mathcal{N}^\mathcal{R}(\tilde{\bm{A}})$ and
    $\bm{\bar{A}}^{t\rightarrow s} =  \mathcal{N}^\mathcal{R}(\tilde{\bm{A}}^\intercal)$ without adding dummy column, keeping the zero rows unchanged when applying $\mathcal{N}^{\mathcal{R}}(\cdot)$. 
    \item Project the source distributions to target by 
    \begin{align*}
        \bar{P}_2(t_q \mid t_p) &= \sum_{i=1}^{|s|} \sum_{j=1}^{|s|} \bm{\bar{A}}^{t\rightarrow s}_{p,i} P_1(s_{j}\mid s_{i}) \bm{\bar{A}}^{s\rightarrow t}_{j,q}
    \end{align*}
    Note that $\bar{P}_2(\cdot\mid\cdot)$ is not guaranteed to be a well formed distribution due to the potential existence of zero rows/columns in $\tilde{\bm{A}}$. 
    \item Normalize $\bar{P}_2(\cdot\mid t_p)$ to $\tilde{P}_2(\cdot\mid t_p)$ for each $p$ separately, ignoring every ``\emph{zero position}'' $p$ that $\bar{P}_2(t_q \mid t_p) = 0$ for all $q$. 
    \item Compute the cross entropy loss between the target parser probability $P_2$ and $\tilde{P}_2$ for all \emph{non-zero positions} $p$. 
\end{enumerate}

We argue that \ddp is equivalent to a weighted sum version to the above approach: that is, there exists a group of weight $(\alpha_1, \ldots, \alpha_{|t|})$ such that the \ddp arc loss $\mathcal{L}_\textit{arc}^{(t)}(P_2, \hat{P}_2) = \sum_{p=1}^{|t|} \alpha_p H(\tilde{P}_2(\cdot\mid t_p), P_2(\cdot\mid t_p))$, where $H(\cdot, \cdot)$ denotes cross entropy, and $H(\cdot, \cdot):= 0$ when the first argument is a ill-formed zero ``distribution''. 

\paragraph{Proof}
First, we note that for all $p = 1, \ldots |t|$ and $i = 1, \ldots, |s|$, 
$$\bm{\bar{A}}^{t\rightarrow s}_{p,i} = \bm{A}^{t\rightarrow s}_{p,i}, $$ 
$$\bm{\bar{A}}^{s\rightarrow t}_{i, p} = \bm{A}^{s\rightarrow t}_{i, p}, $$
as adding dummy column does not affect the normalization result for non-dummy positions. 

Therefore, 
\begin{align*}
    \hat{P}_2(t_q \mid t_p) =& \sum_{i=1}^{|s|+1} \sum_{j=1}^{|s|} \bm{A}^{t\rightarrow s}_{p,i} P_1(s_{j}\mid s_{i}) \bm{A}^{s\rightarrow t}_{j,q} \\
    =& \sum_{i=1}^{|s|} \sum_{j=1}^{|s|} \bm{A}^{t\rightarrow s}_{p,i} P_1(s_{j}\mid s_{i}) \bm{A}^{s\rightarrow t}_{j,q} + \\
    & \sum_{j=1}^{|s|} \bm{A}^{t\rightarrow s}_{p,|s|+1} P_1(s_{j}\mid s_{|s|+1}) \bm{A}^{s\rightarrow t}_{j,q} \\
    =& \sum_{i=1}^{|s|} \sum_{j=1}^{|s|} \bm{A}^{t\rightarrow s}_{p,i} P_1(s_{j}\mid s_{i}) \bm{A}^{s\rightarrow t}_{j,q} \\
    =& \sum_{i=1}^{|s|} \sum_{j=1}^{|s|} \bm{\bar{A}}^{t\rightarrow s}_{p,i} P_1(s_{j}\mid s_{i}) \bm{\bar{A}}^{s\rightarrow t}_{j,q} \\
    =& \bar{P}_2 (t_q \mid t_p)
\end{align*}
That is, $\tilde{P}_2(\cdot\mid t_p)$ is can be also calculated by normalization of $\hat{P}_2 (t_q \mid t_p)$, where $q = 1, \ldots, |t|$.\footnote{We may here intuitively view that the dummy position $P(t_{|t|+1}\mid t_p)$ absorbs some original probability corresponding to unaligned words. }
Therefore, for any $p=1, \ldots, |t|$, there exists $\alpha_p$ such that $ \hat{P}_2(\cdot\mid t_p) = \alpha_p \tilde{P}_2(\cdot\mid t_p)$. 

By definition, 
\begin{align*}
    &\mathcal{L}_\textit{arc}^{(t)}(P_2, \hat{P}_2) \\
    = &-\sum_{p=1}^{|t|}\sum_{q=1}^{|t|}\hat{P}_2(t_q \mid t_p)\log P_2(t_q \mid t_p) \\ 
    = &-\sum_{p=1}^{|t|}\sum_{q=1}^{|t|}\alpha_p \tilde{P}_2(t_q \mid t_p)\log P_2(t_q \mid t_p) \\
    = &\sum_{p=1}^{|t|} \alpha_p H(\tilde{P}_2(\cdot\mid t_p), P_2(\cdot\mid t_p)).  
\end{align*}

\begin{figure}[t!]
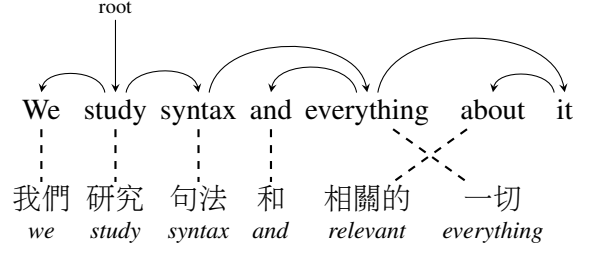
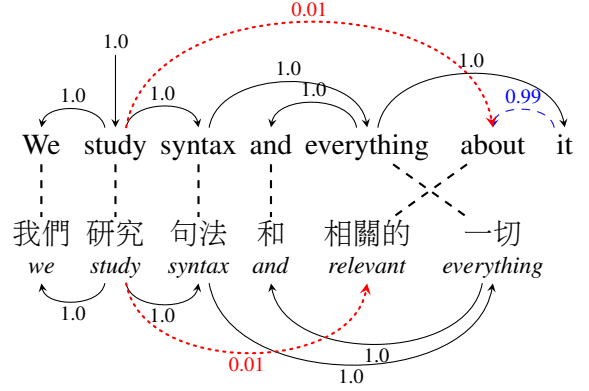
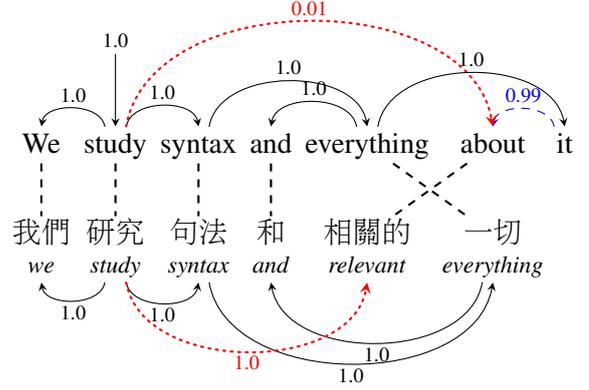

    \centering
    \begin{subfigure}[b]{0.45\textwidth}
    \begin{dependency}[arc edge, arc angle=80, text only label, label style={above}]
    \begin{deptext}[column sep=.0cm]
    We \& study \& syntax \& and \& everything \& about \& it \\[0.8cm]
    我們 \& 研究 \& 句法 \& 和 \& 相關的 \& 一切  \\
    \small \textit{we} \& \small \textit{study} \& \small \textit{syntax} \& \small \textit{and} \& \small \textit{relevant} \& \small \textit{everything} \\
    \end{deptext} 
    \depedge{2}{1}{}
    \depedge{2}{3}{}
    \depedge{5}{4}{}
    \depedge{3}{5}{}
    \depedge{7}{6}{}
    \depedge{5}{7}{}
    \deproot[edge unit distance=0.3cm]{2}{root}
    \draw [-, thick, dashed] (\wordref{1}{1}) -- (\wordref{2}{1});
    \draw [-, thick, dashed] (\wordref{1}{2}) -- (\wordref{2}{2});
    \draw [-, thick, dashed] (\wordref{1}{3}) -- (\wordref{2}{3});
    \draw [-, thick, dashed] (\wordref{1}{4}) -- (\wordref{2}{4});
    \draw [-, thick, dashed] (\wordref{1}{5}) -- (\wordref{2}{6});
    \draw [-, thick, dashed] (\wordref{1}{6}) -- (\wordref{2}{5});
    \end{dependency}  \\[-0.2cm]
    \caption{Ground-truth unlabeled parse tree and alignment.}
    \end{subfigure}
    
    \begin{subfigure}[b]{0.45\textwidth}
    \begin{dependency}[arc edge, arc angle=80, text only label, label style={above}]
    \begin{deptext}[column sep=.0cm]
    We \& study \& syntax \& and \& everything \& about \& it \\[0.8cm]
    我們 \& 研究 \& 句法 \& 和 \& 相關的 \& 一切  \\
    \small \textit{we} \& \small \textit{study} \& \small \textit{syntax} \& \small \textit{and} \& \small \textit{relevant} \& \small \textit{everything} \\
    \end{deptext} 
    \depedge{2}{1}{1.0}
    \depedge{2}{3}{1.0}
    \depedge{5}{4}{1.0}
    \depedge{3}{5}{1.0}
    \depedge[blue, dashed]{7}{6}{\textcolor{blue}{0.99}}
    \depedge[red, dotted, thick]{2}{6}{\textcolor{red}{0.01}}
    \depedge{5}{7}{1.0}
    \deproot[edge unit distance=0.3cm]{2}{1.0}
    \draw [-, thick, dashed] (\wordref{1}{1}) -- (\wordref{2}{1});
    \draw [-, thick, dashed] (\wordref{1}{2}) -- (\wordref{2}{2});
    \draw [-, thick, dashed] (\wordref{1}{3}) -- (\wordref{2}{3});
    \draw [-, thick, dashed] (\wordref{1}{4}) -- (\wordref{2}{4});
    \draw [-, thick, dashed] (\wordref{1}{5}) -- (\wordref{2}{6});
    \draw [-, thick, dashed] (\wordref{1}{6}) -- (\wordref{2}{5});
    \depedge[label style=below, edge below]{2}{1}{1.0}
    \depedge[label style=below, edge below]{2}{3}{1.0}
    \depedge[label style=below, edge below]{6}{4}{1.0}
    \depedge[label style=below, edge below]{3}{6}{1.0}
    \depedge[label style=below, edge below, red, dotted, thick]{2}{5}{\textcolor{red}{0.01}}
    \end{dependency}  \\[-0.2cm]
    \caption{Projection by \ddp of the arc distributions predicted by a neural parser. Label denotes edge probability. }
    \label{fig:not-normalize}
    \end{subfigure}
    
    \begin{subfigure}[b]{0.45\textwidth}
    \begin{dependency}[arc edge, arc angle=80, text only label, label style={above}]
    \begin{deptext}[column sep=.0cm]
    We \& study \& syntax \& and \& everything \& about \& it \\[0.8cm]
    我們 \& 研究 \& 句法 \& 和 \& 相關的 \& 一切  \\
    \small \textit{we} \& \small \textit{study} \& \small \textit{syntax} \& \small \textit{and} \& \small \textit{relevant} \& \small \textit{everything} \\
    \end{deptext} 
    \depedge{2}{1}{1.0}
    \depedge{2}{3}{1.0}
    \depedge{5}{4}{1.0}
    \depedge{3}{5}{1.0}
    \depedge[blue, dashed]{7}{6}{\textcolor{blue}{0.99}}
    \depedge[red, dotted, thick]{2}{6}{\textcolor{red}{0.01}}
    \depedge{5}{7}{1.0}
    \deproot[edge unit distance=0.3cm]{2}{1.0}
    \draw [-, thick, dashed] (\wordref{1}{1}) -- (\wordref{2}{1});
    \draw [-, thick, dashed] (\wordref{1}{2}) -- (\wordref{2}{2});
    \draw [-, thick, dashed] (\wordref{1}{3}) -- (\wordref{2}{3});
    \draw [-, thick, dashed] (\wordref{1}{4}) -- (\wordref{2}{4});
    \draw [-, thick, dashed] (\wordref{1}{5}) -- (\wordref{2}{6});
    \draw [-, thick, dashed] (\wordref{1}{6}) -- (\wordref{2}{5});
    \depedge[label style=below, edge below]{2}{1}{1.0}
    \depedge[label style=below, edge below]{2}{3}{1.0}
    \depedge[label style=below, edge below]{6}{4}{1.0}
    \depedge[label style=below, edge below]{3}{6}{1.0}
    \depedge[label style=below, edge below, red, dotted, thick]{2}{5}{\textcolor{red}{1.0}}
    \end{dependency}  \\[-0.2cm]
    \caption{Projection by the alternative algorithm $\mathcal{A}$ of the arc distributions. Label denotes edge probability. }
    \label{fig:normalize}
    \end{subfigure}
    
    \caption{Intuition on the reason that we do not apply normalization in \ddp as described in the alternative algorithm $\mathcal{A}$. }
    \label{fig:normalize-or-not}
\end{figure}

We use toy example (Figure~\ref{fig:normalize-or-not}) to show the intuition for using \ddp instead of the alternative approach $\mathcal{A}$. 
It is common for neural network--based parsers generate a very low non-zero arc probability for a random word pair with no direct dependency relation, e.g., (study $\rightarrow$ about): normalization of arc probability, may significantly enlarge the noise level when the correct arc (it $\rightarrow$ about in this case) is not projected due to alignment mismatch, weighting undesirable target language arcs (e.g., 研究$\rightarrow$ 相關的 ;Figure~\ref{fig:normalize}) as much as those with high quality in the training loss; in contrast, while \ddp (Figure~\ref{fig:not-normalize}) may also introduce such noise, the corresponding weight remains in the same scale as the probability predicted by the neural parser. 

\section{Implementation Details of the Bi-Affine Dependency Parser}
\label{sec:biaffine-parser}
Given a sentence $s$, we extract the subword representations by a pretrained multilingual contextualized representation model (XLM-R or CRISS), and take endpoint concatenation of corresponding subwords representations as word representations, yielding contextualized word features $\bm{V}\in \mathbb{R}^{|s| \times d}$, where $|s|$ denotes the number of words in $s$, and 
$d$ denotes the dimensionality of the extracted features. 
We further perform non-linear transformation on the features with multi-layer perceptrons (MLPs) with ReLU activation and a long short-term memory module \citep[LSTM;][]{hochreiter1997long}, to obtain head and dependent features $\bm{H}$ and $\bm{D}$:\footnote{We find that the LSTM module is important, removing it will result in 1-2 points drop in terms of both UAS and LAS, in the supervised training settings for English.}
\begin{align}
    \tilde{\bm{V}} &= \text{LSTM}(\text{MLP}_\textit{feature}(\bm{V})) \nonumber \\
    \bm{H} &= \text{MLP}_\textit{head}(\tilde{\bm V}) \nonumber \\
    \bm{D} &= \text{MLP}_\textit{dependent}(\tilde{\bm V}). \nonumber 
\end{align}

\section{Cross-Lingual Transfer Results on Individual Languages}
\label{sec:invididual-plots}
\begin{figure*}[t]
    \includegraphics[width=0.24\textwidth]{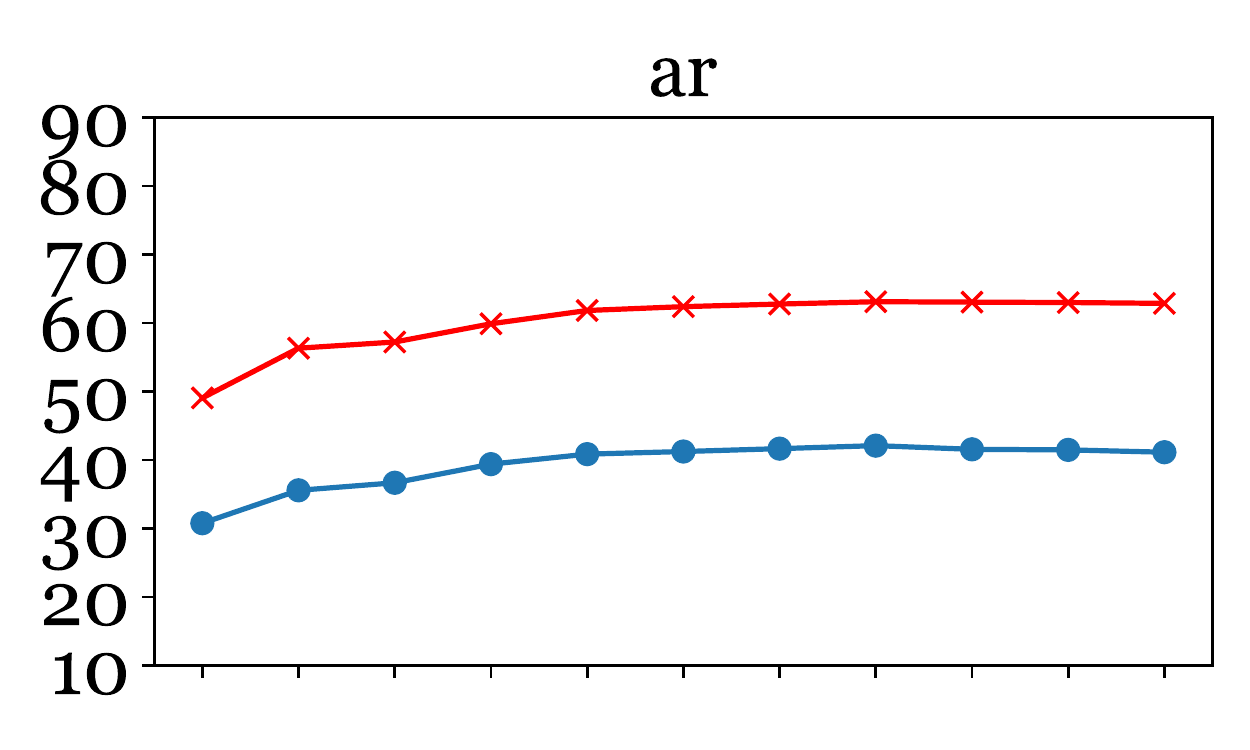}
    \includegraphics[width=0.24\textwidth]{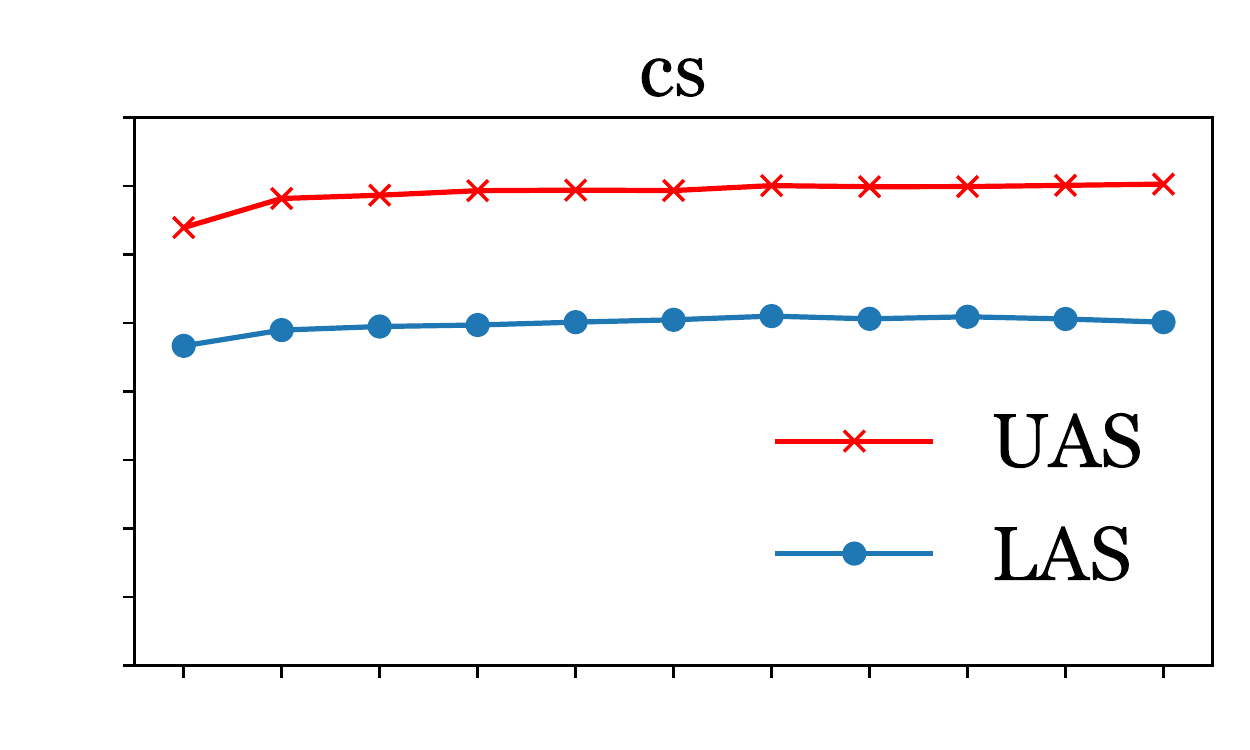} \\
    \includegraphics[width=0.24\textwidth]{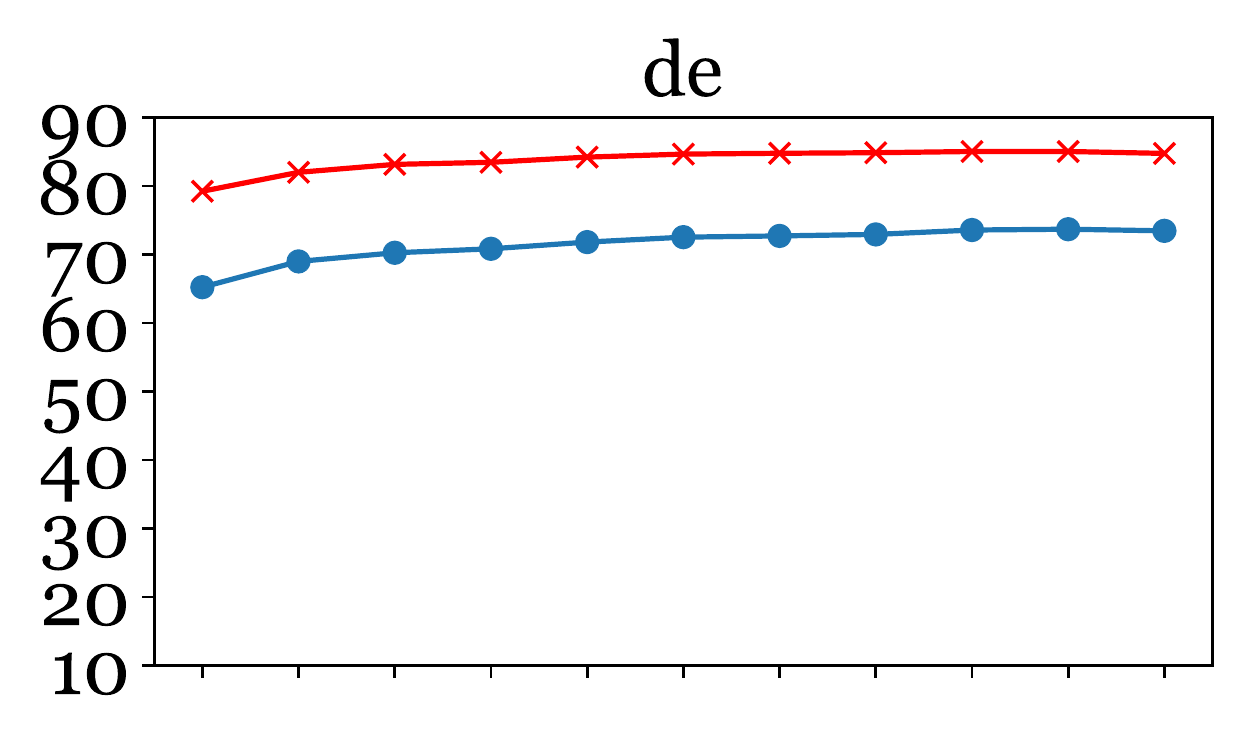}
    \includegraphics[width=0.24\textwidth]{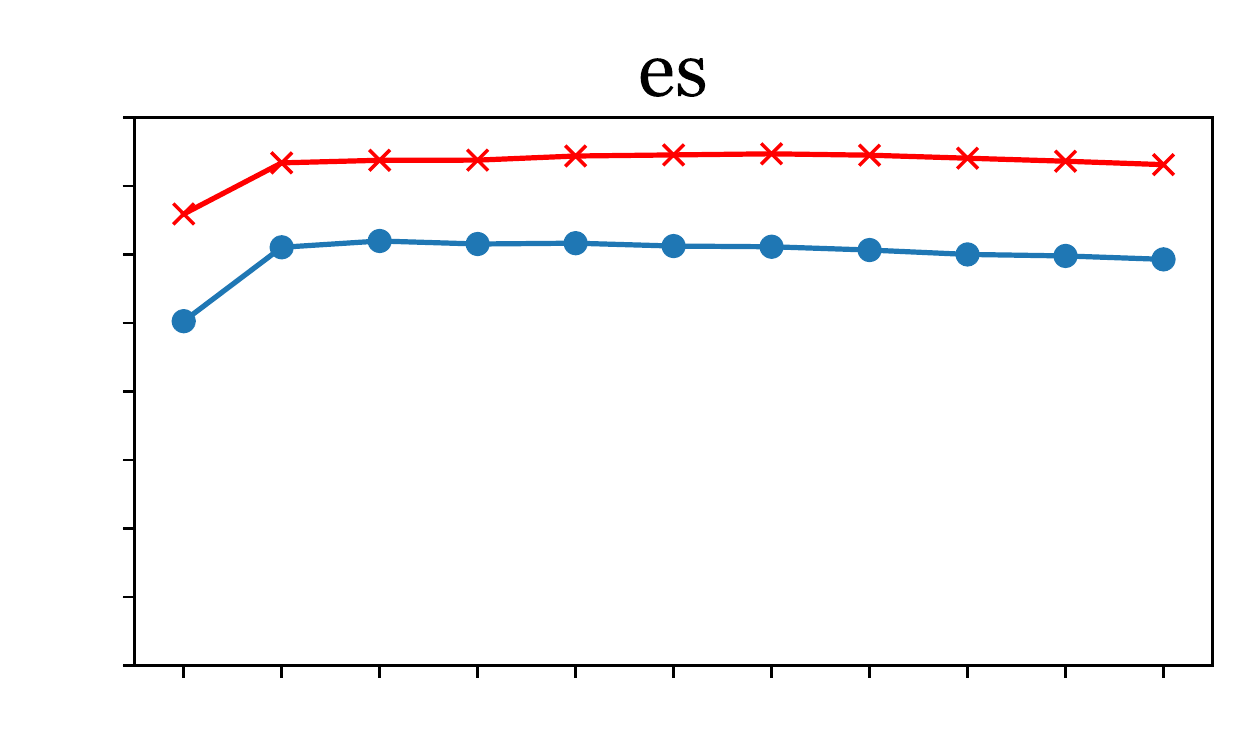} 
    \includegraphics[width=0.24\textwidth]{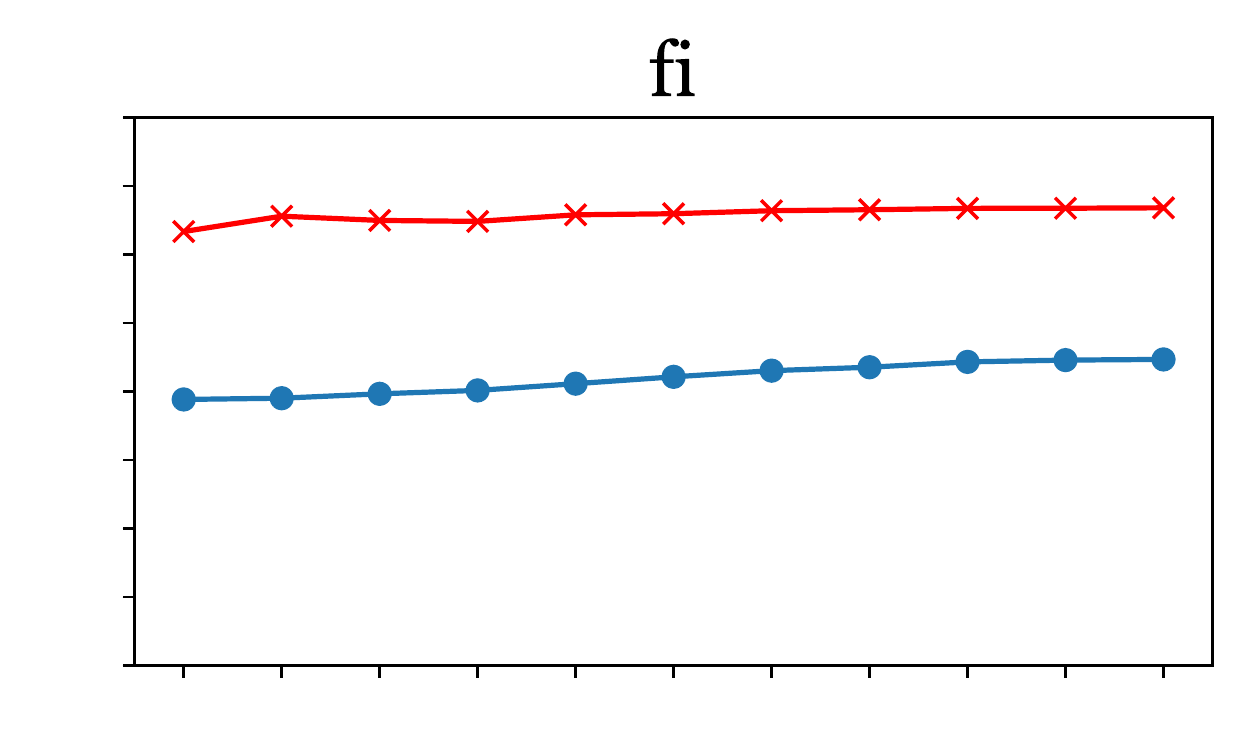}
    \includegraphics[width=0.24\textwidth]{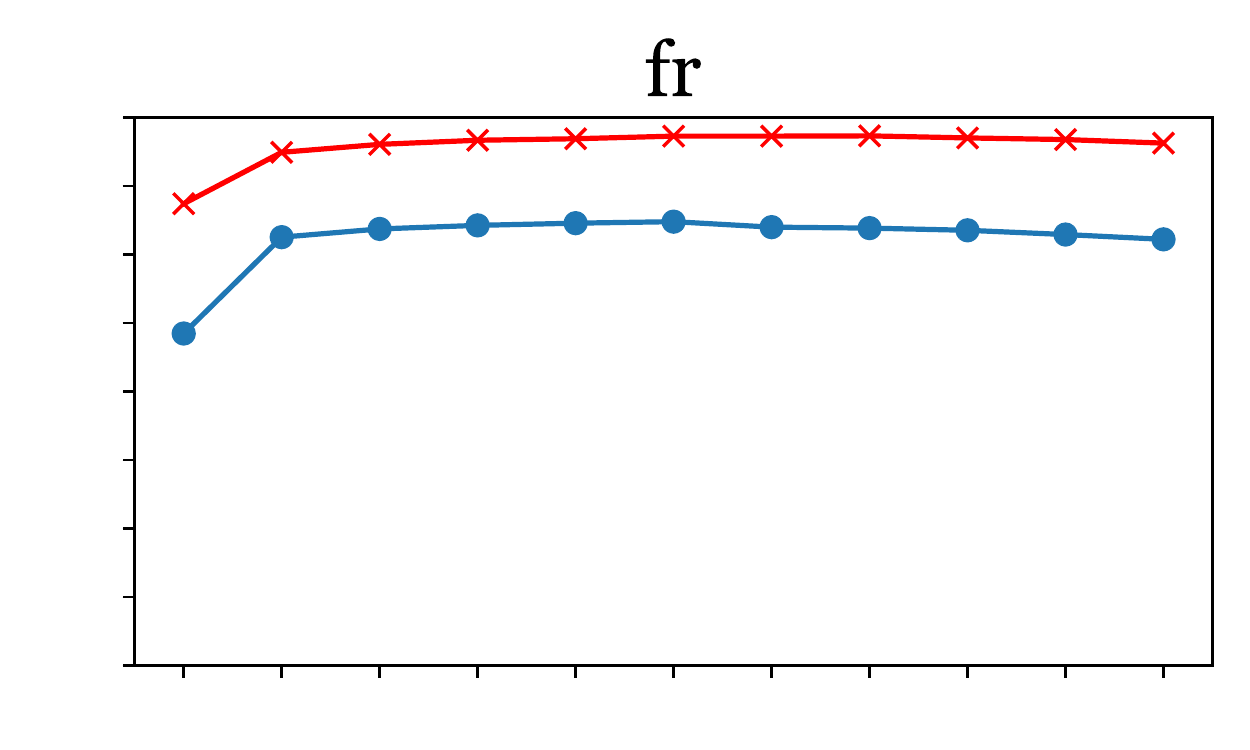} \\
    \includegraphics[width=0.24\textwidth]{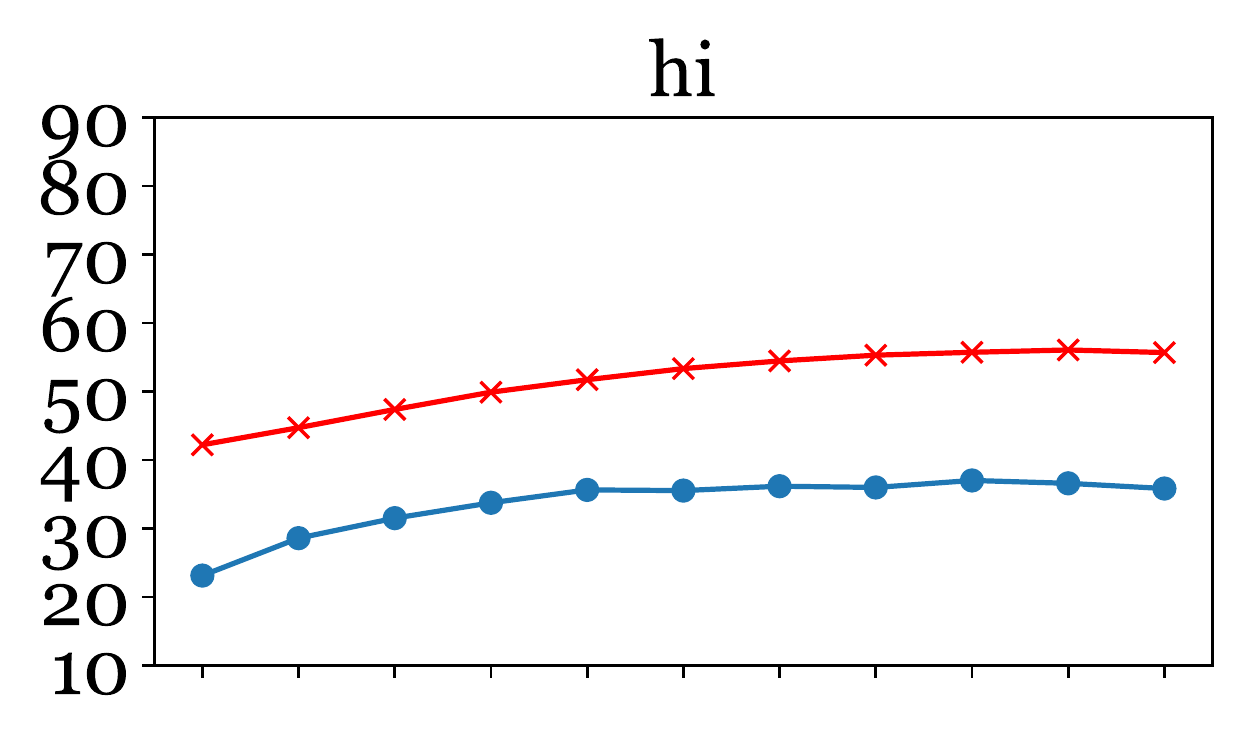}
    \includegraphics[width=0.24\textwidth]{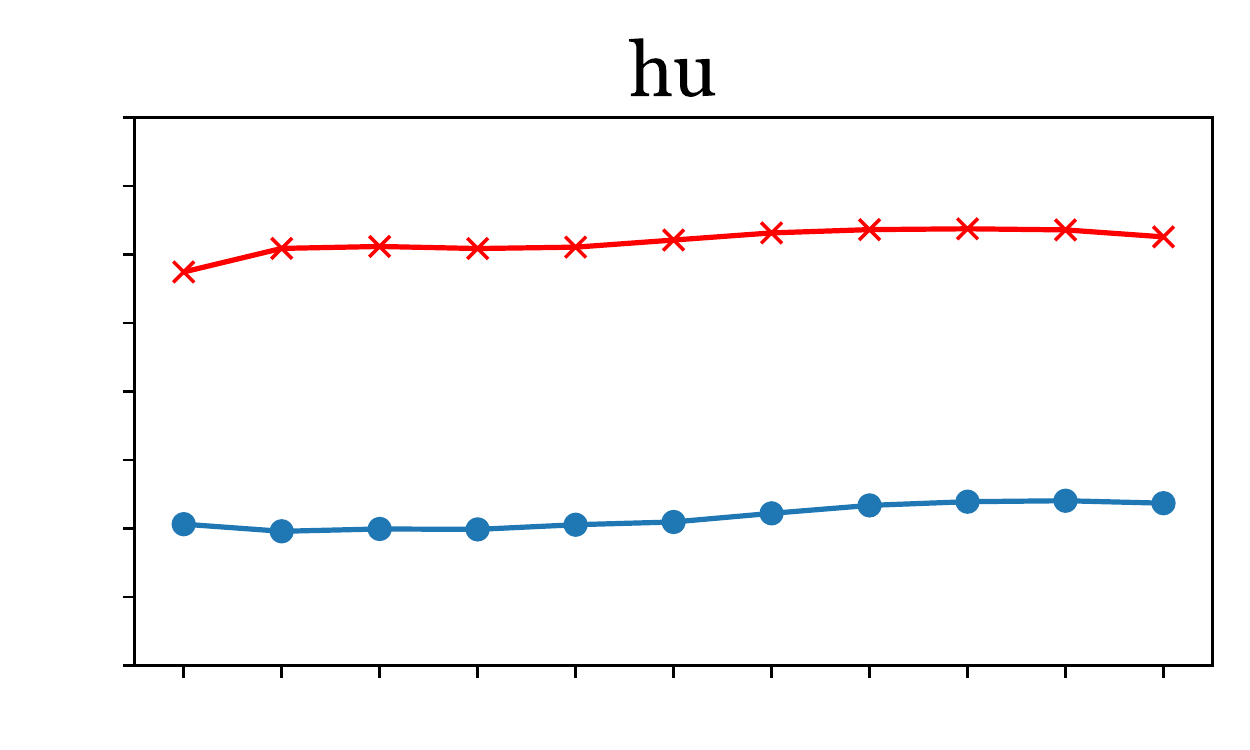} 
    \includegraphics[width=0.24\textwidth]{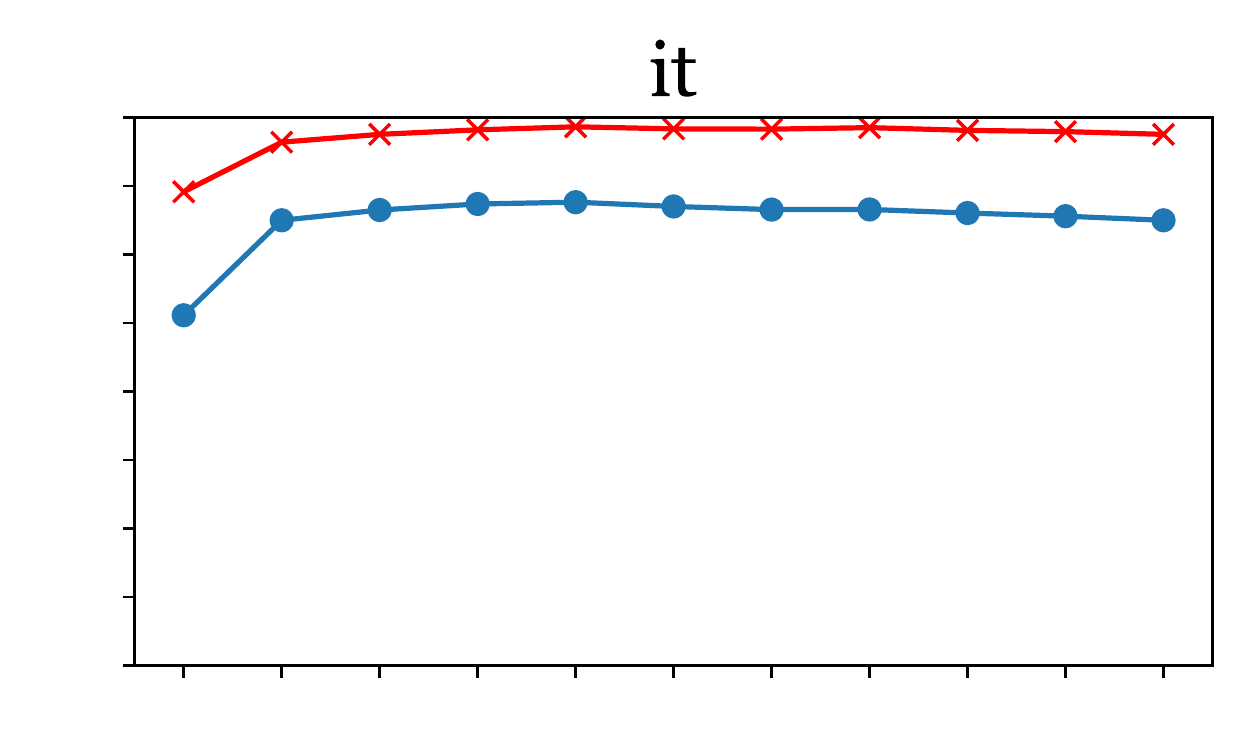}
    \includegraphics[width=0.24\textwidth]{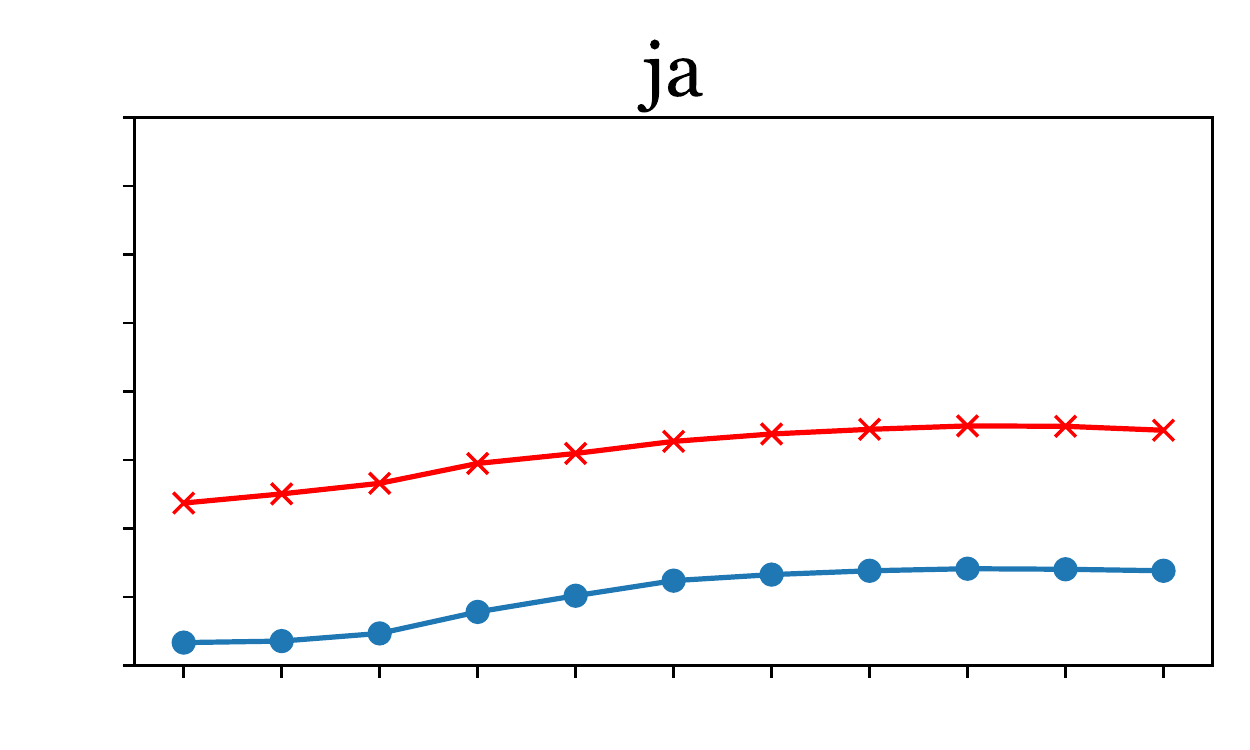} \\
    \includegraphics[width=0.24\textwidth]{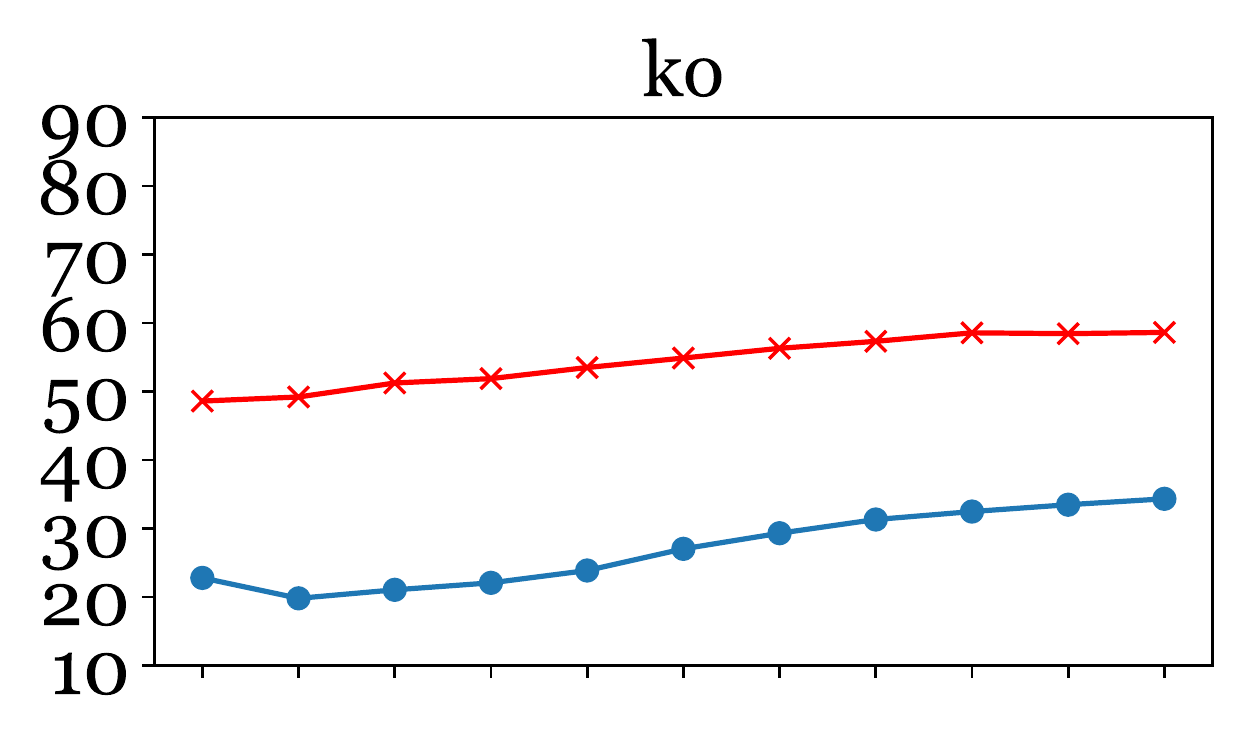}
    \includegraphics[width=0.24\textwidth]{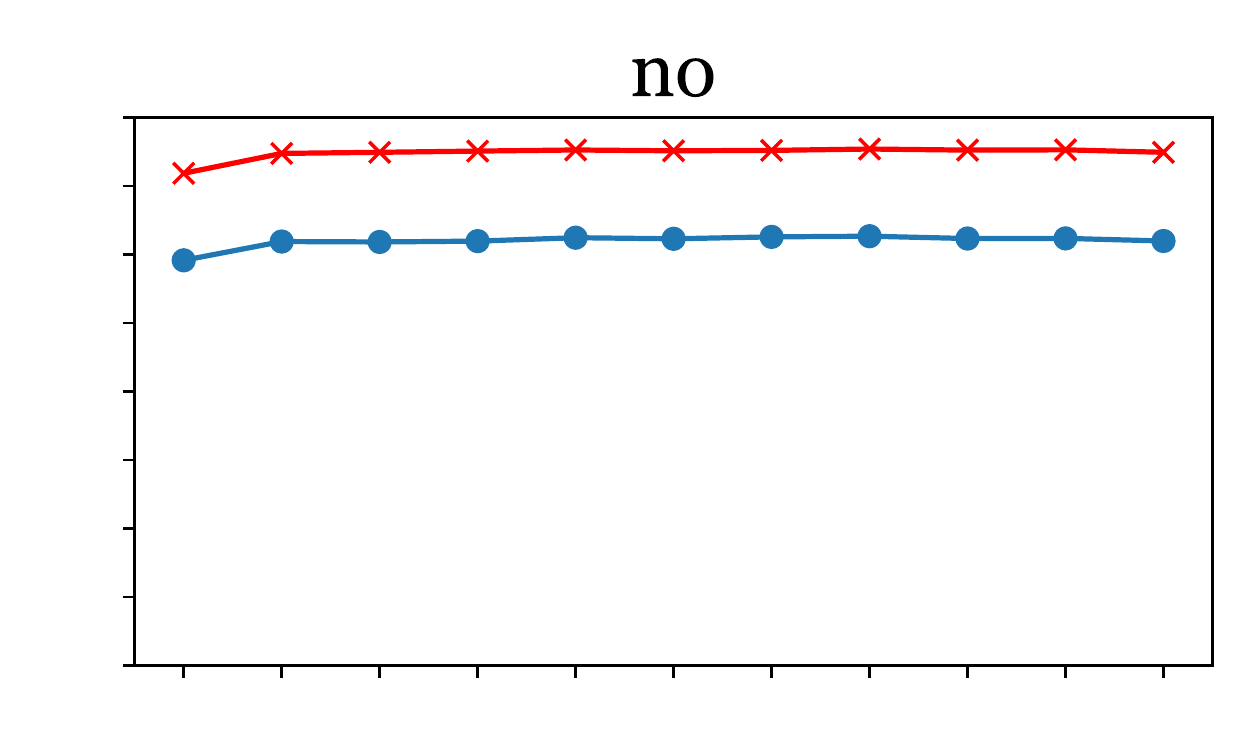} 
    \includegraphics[width=0.24\textwidth]{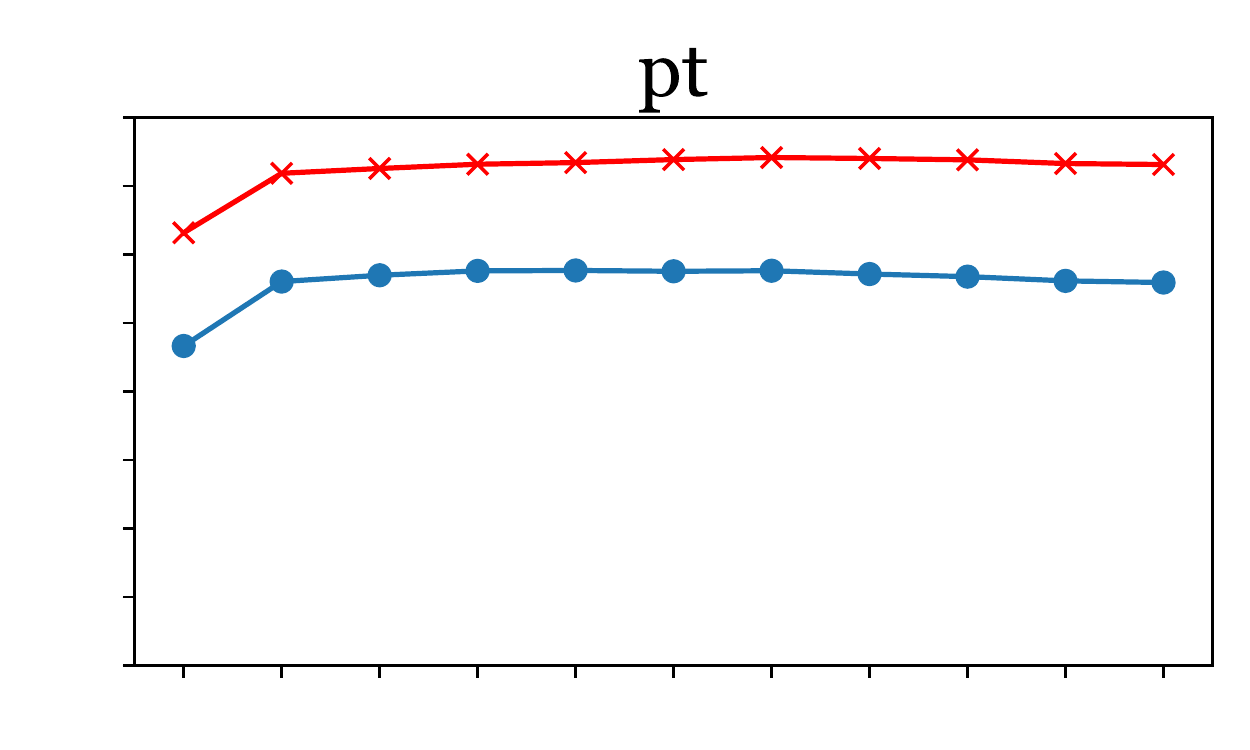}
    \includegraphics[width=0.24\textwidth]{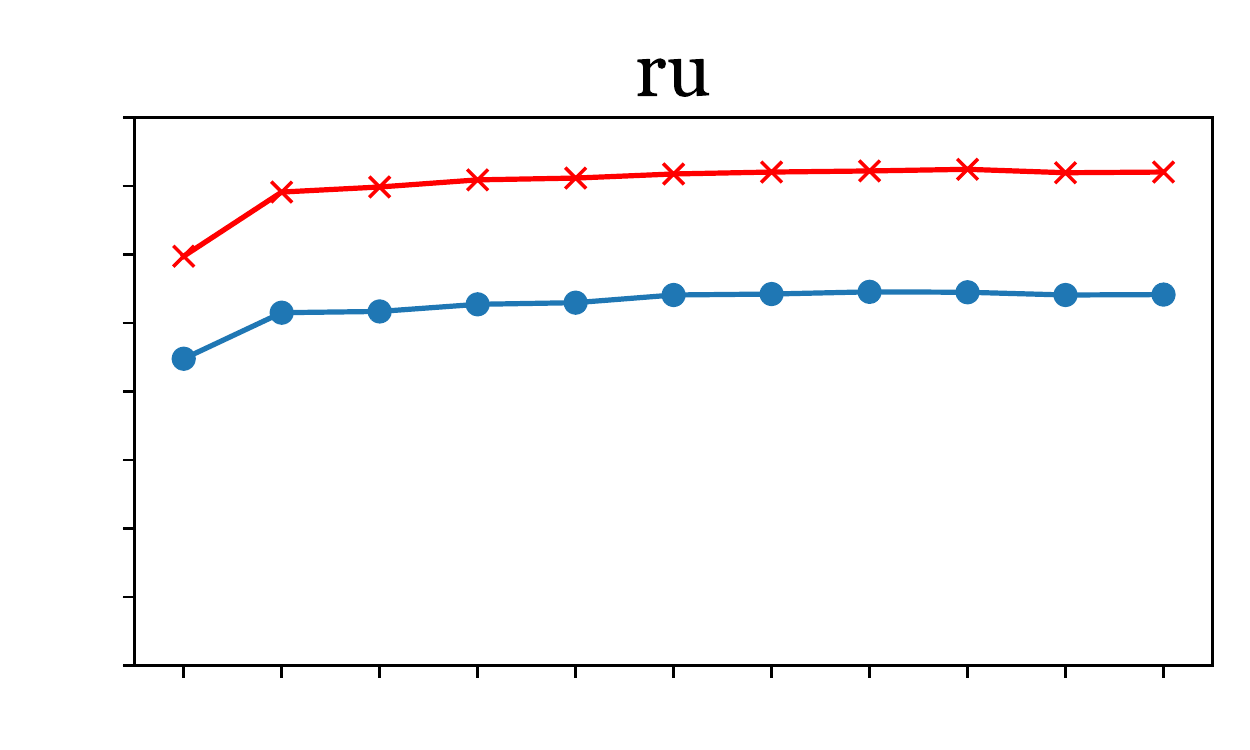} \\
    \includegraphics[width=0.24\textwidth]{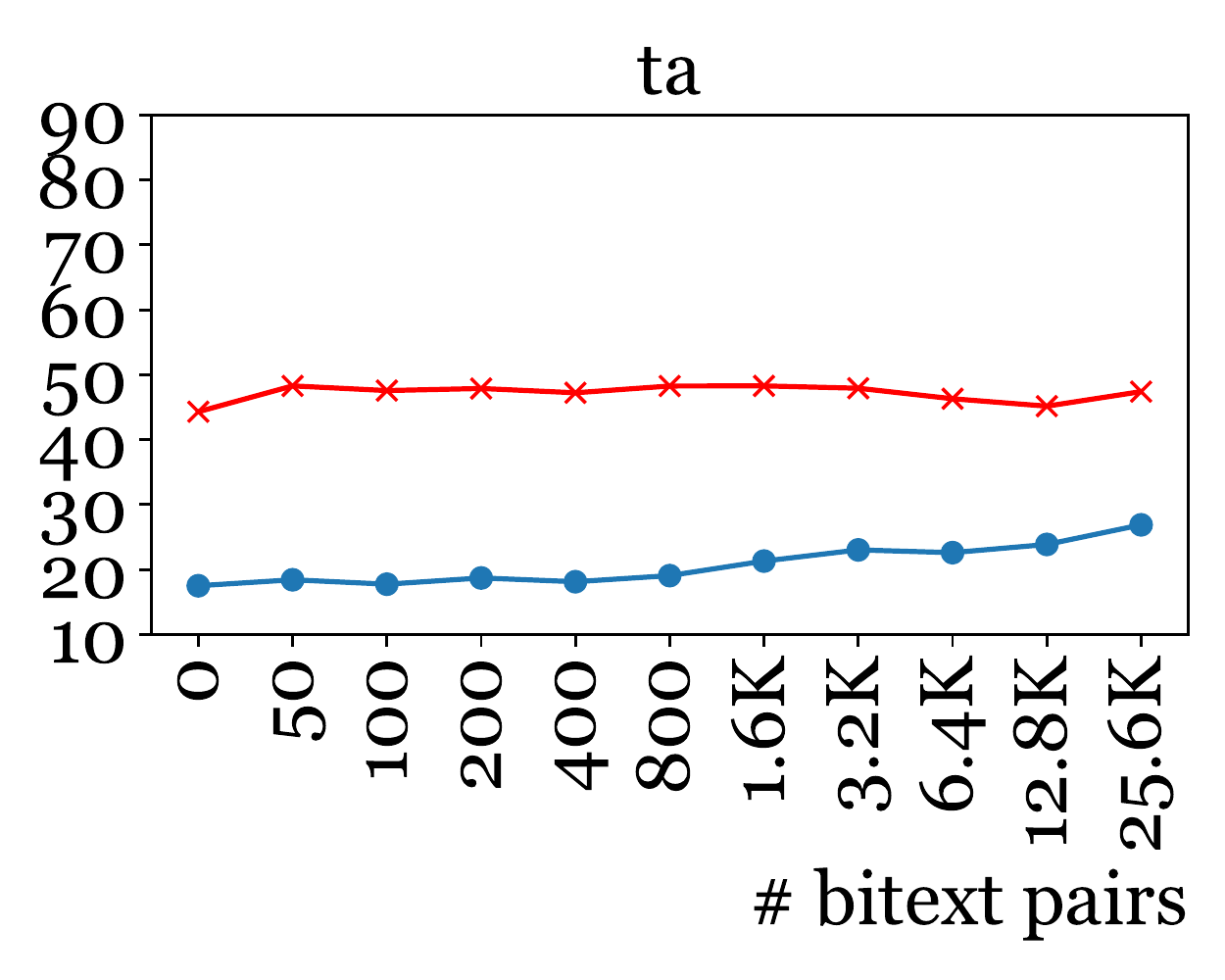}
    \includegraphics[width=0.24\textwidth]{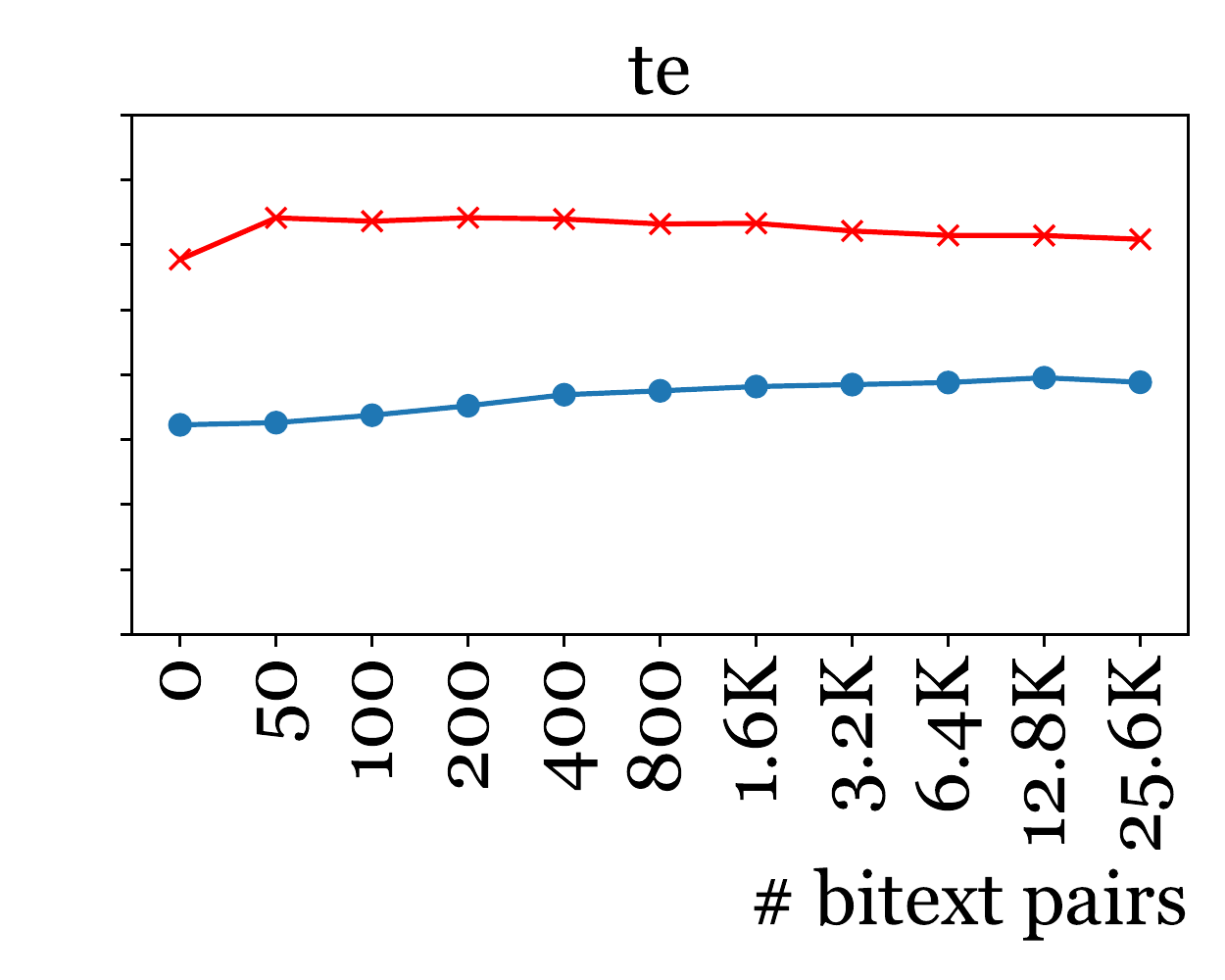} 
    \includegraphics[width=0.24\textwidth]{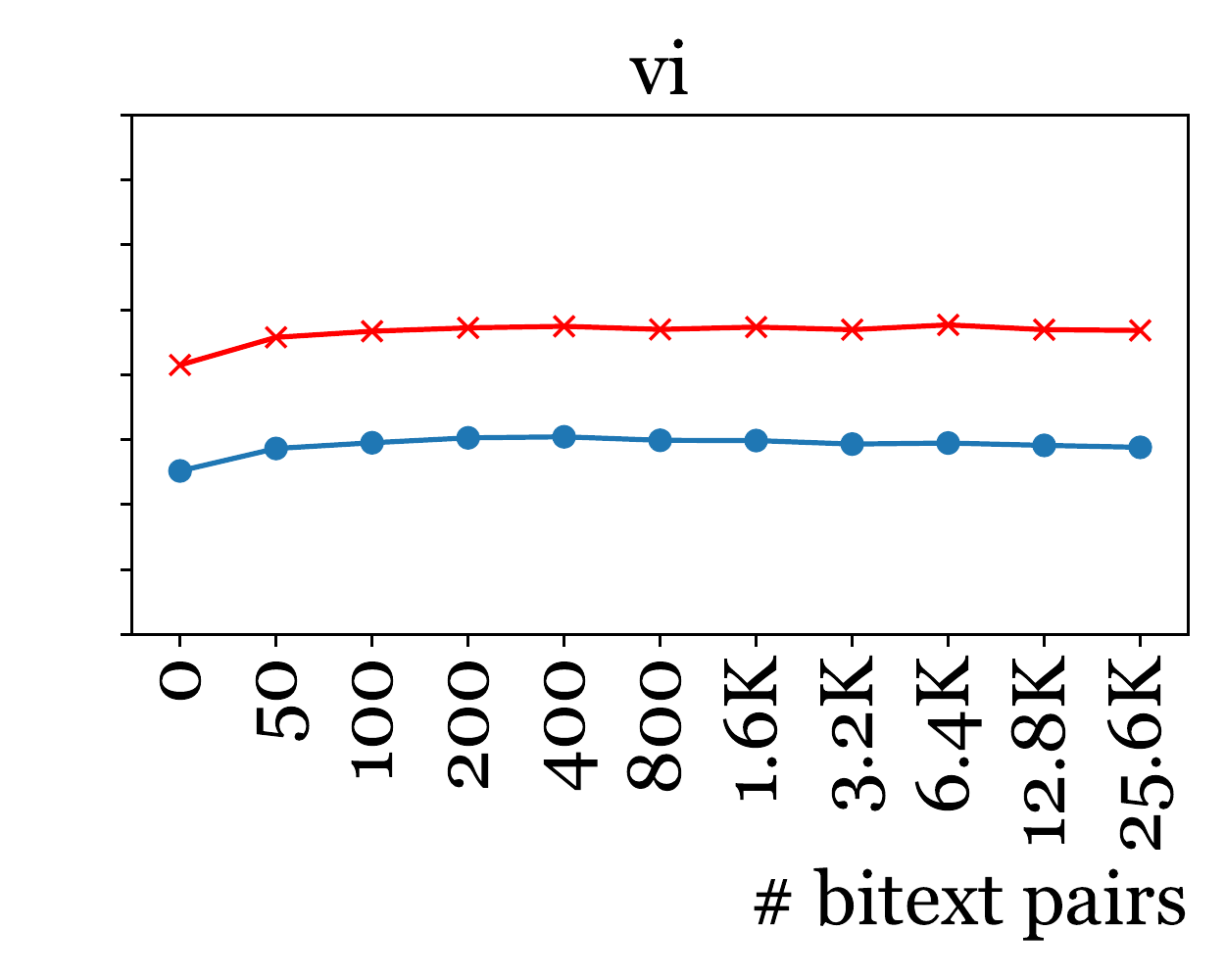}
    \includegraphics[width=0.24\textwidth]{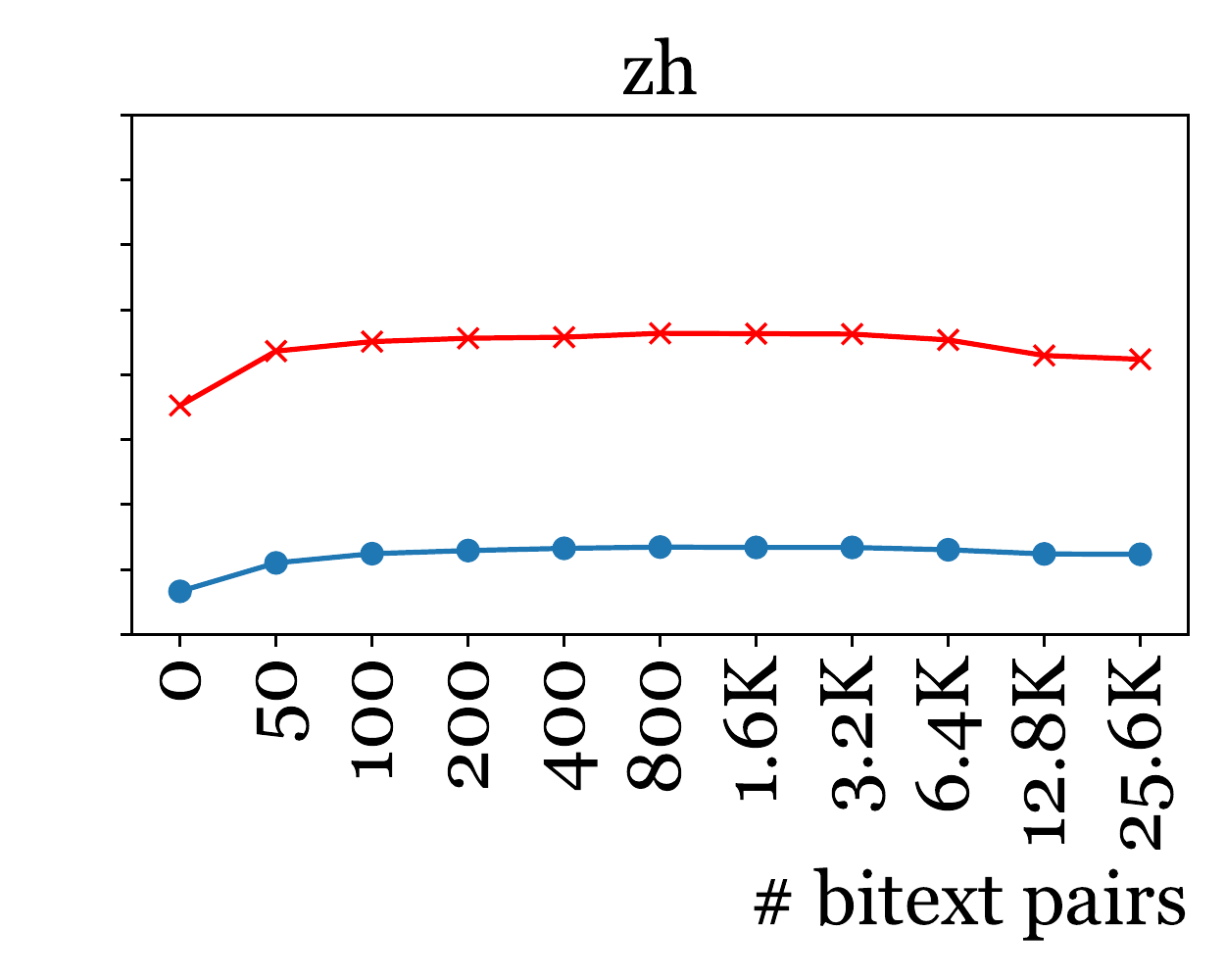} 
    \caption{LAS and UAS on the Universal Dependencies v2.2 standard development sets (best viewed in color), where the $x$-axis represents the number of bitext pairs used, and $y$-axis corresponds to attachment scores. All numbers are averaged across 5 runs with different random seeds and different sets of bitext if applicable. }
    \label{fig:data-efficiency-individual-lang}
\end{figure*}

We present the \ddp zero-shot cross-lingual dependency parsing performance for each individual language with respect to the numbers of bitext pairs (Figure~\ref{fig:data-efficiency-individual-lang}). 
\ddp with supervised bitext outperforms the direct transfer baseline (using 0 pair of bitext) for all languages. 
For most languages, \ddp starts improves over direct transfer with only 50 pairs of bitext. 

\section{Ablation Study in UAS}
\begin{figure*}[t]
\includegraphics[width=0.27\textwidth]{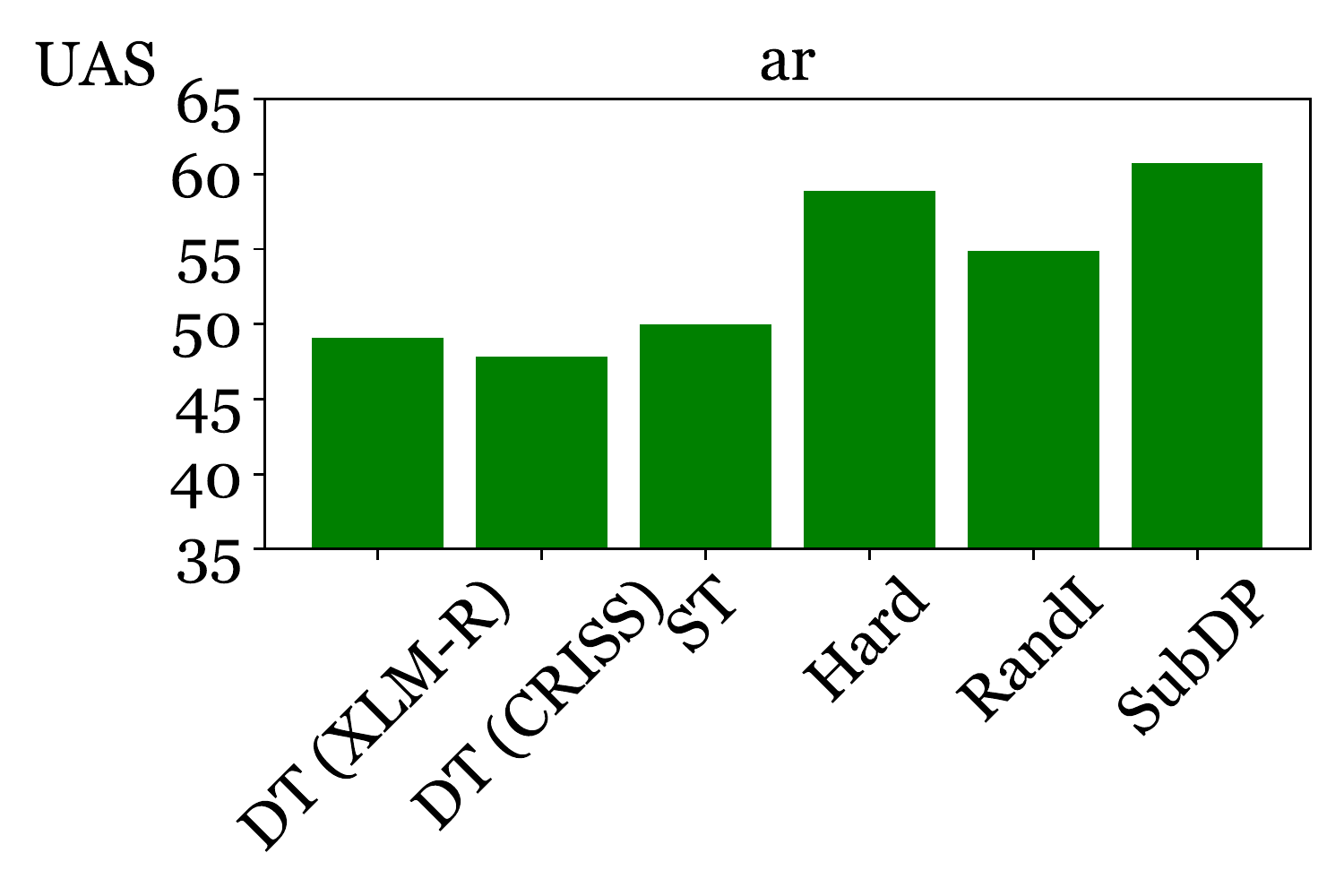} \hspace{-25pt}
\includegraphics[width=0.27\textwidth]{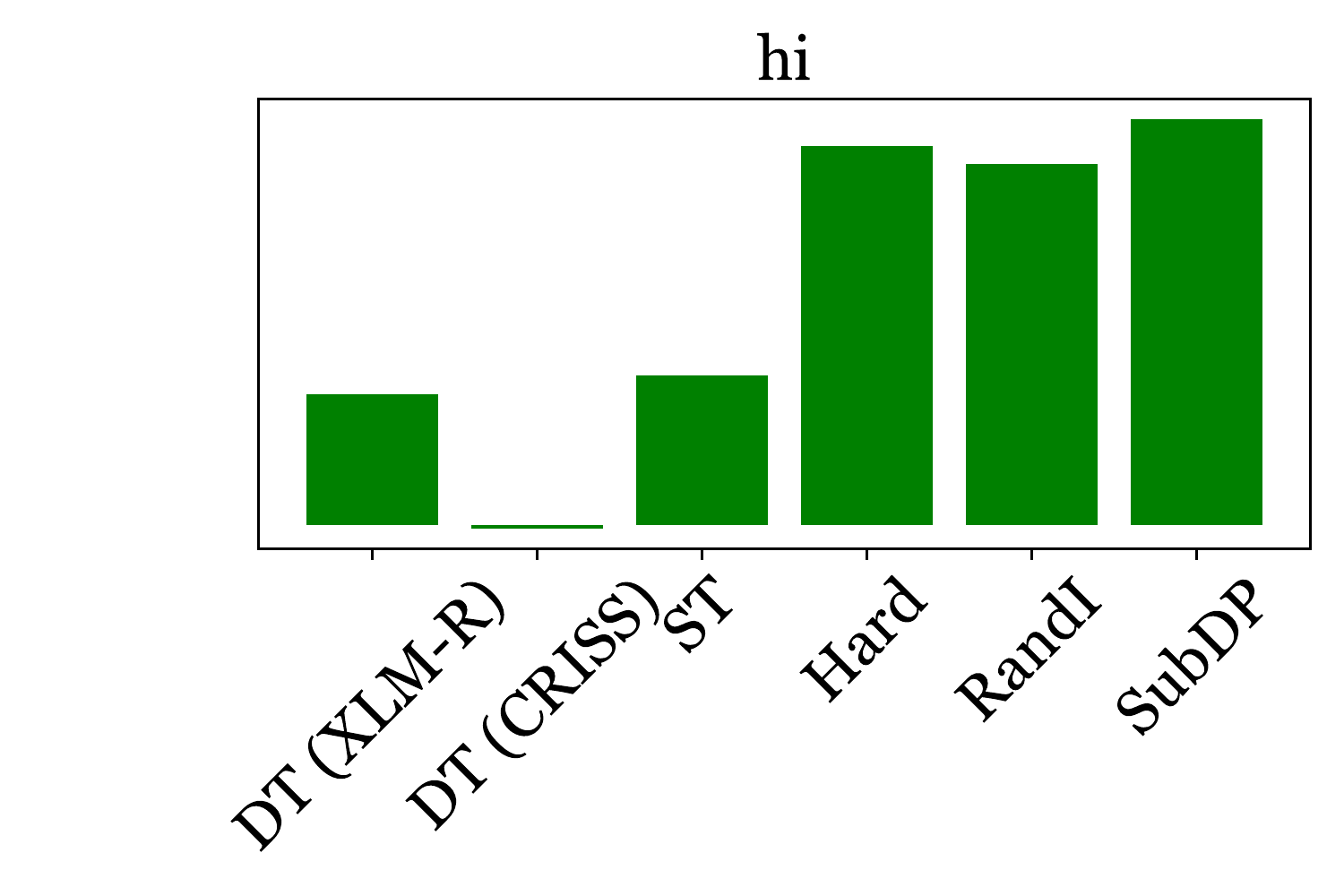} \hspace{-25pt}
\includegraphics[width=0.27\textwidth]{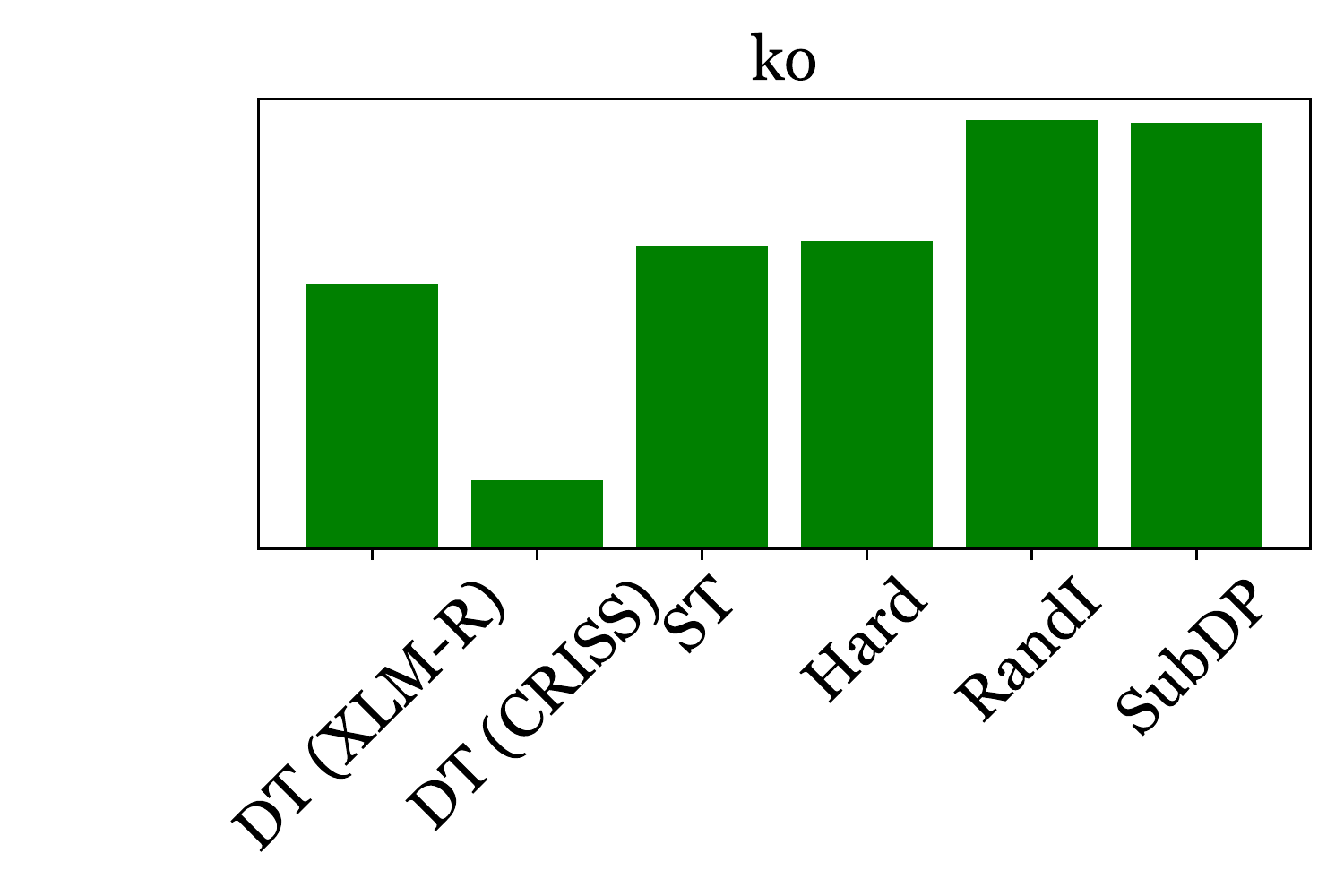} \hspace{-25pt}
\includegraphics[width=0.27\textwidth]{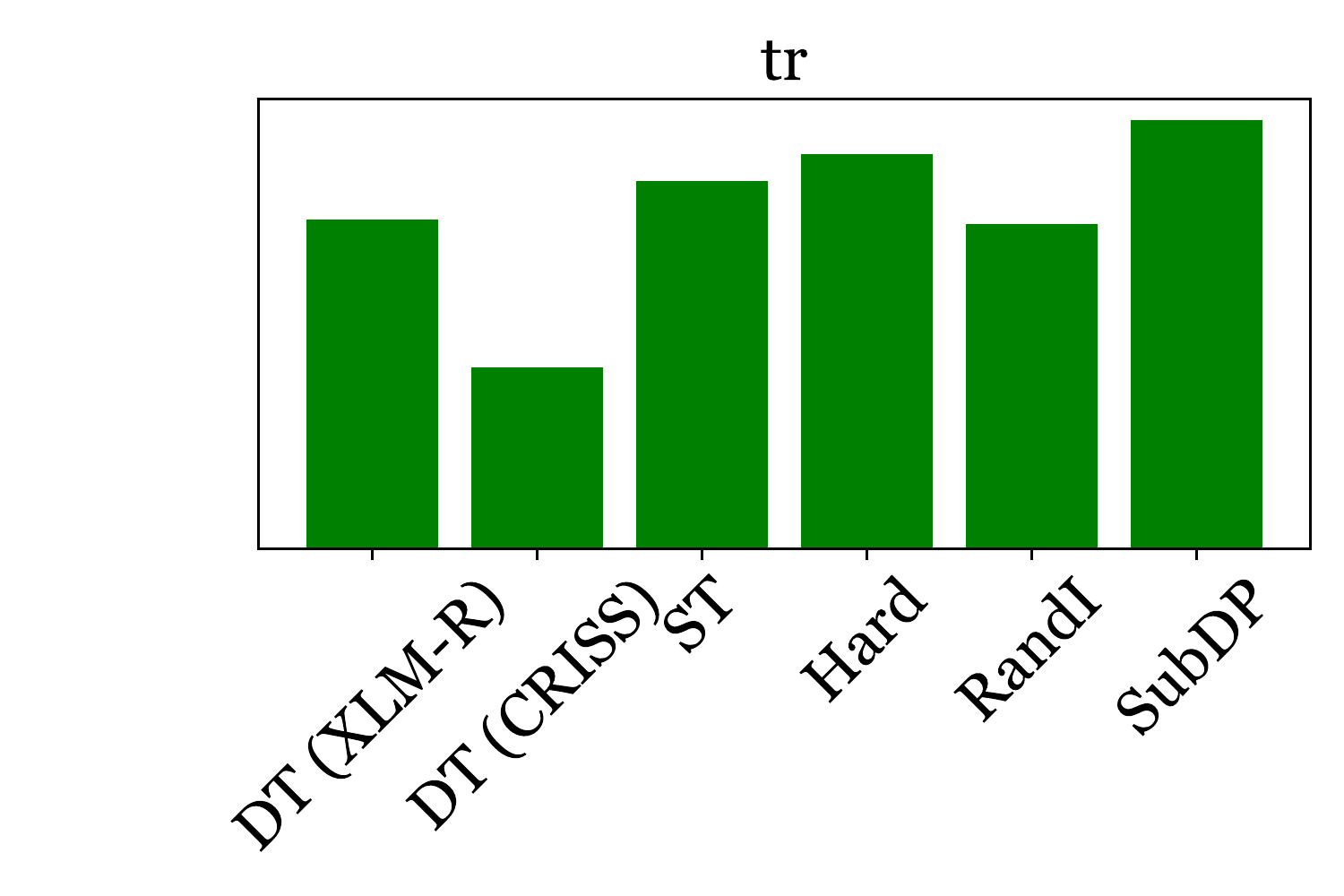} \\
\includegraphics[width=0.27\textwidth]{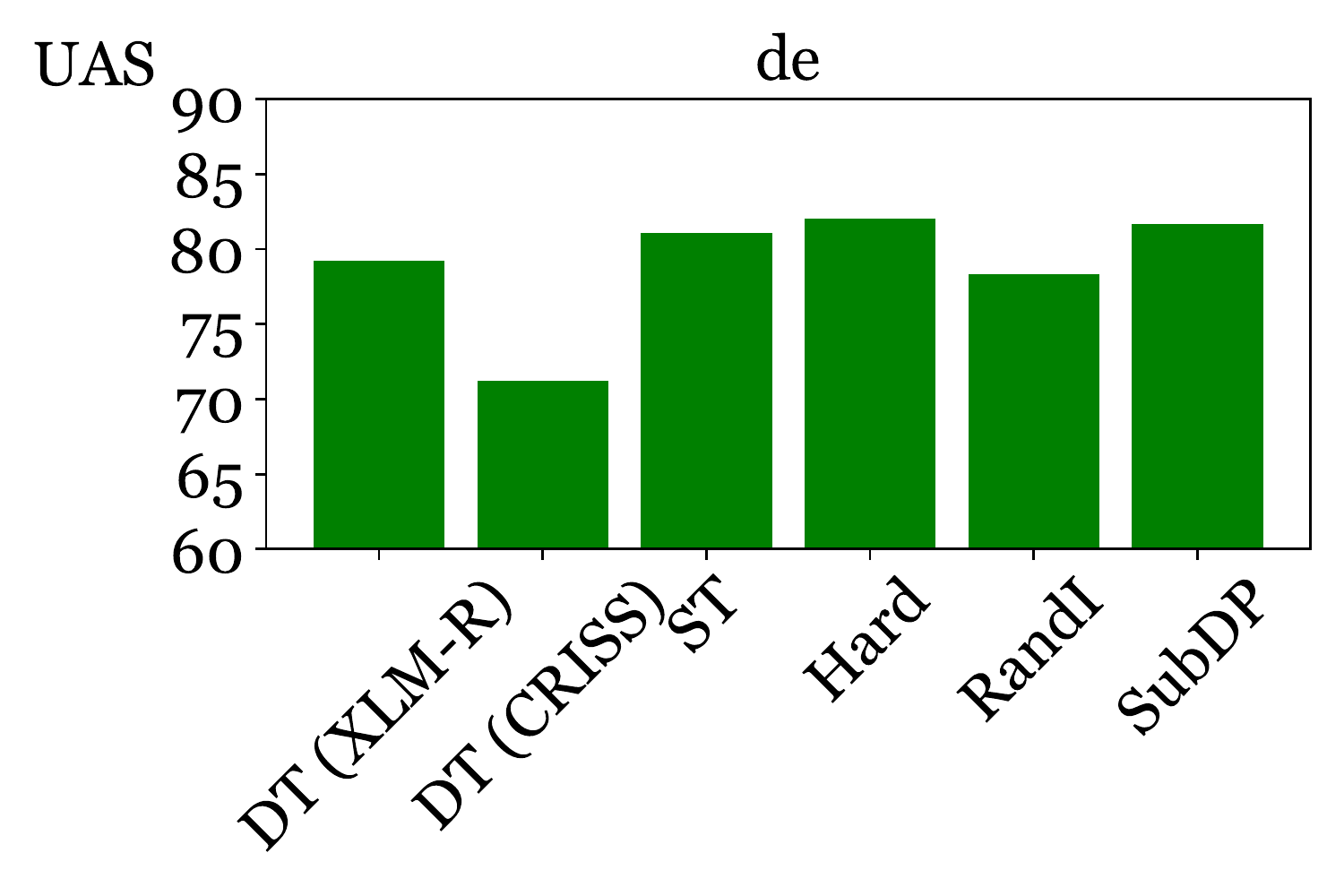} \hspace{-25pt}
\includegraphics[width=0.27\textwidth]{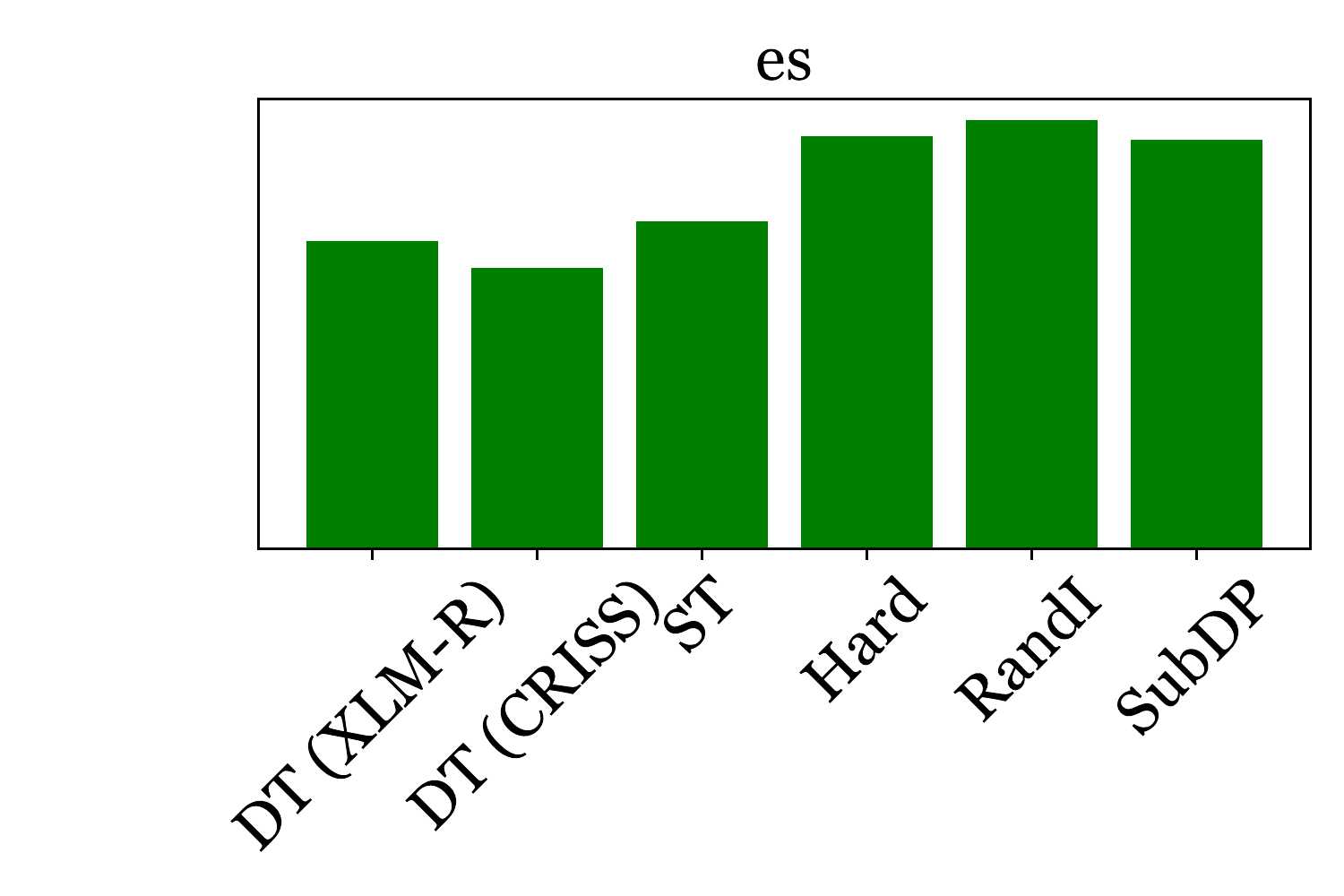} \hspace{-25pt}
\includegraphics[width=0.27\textwidth]{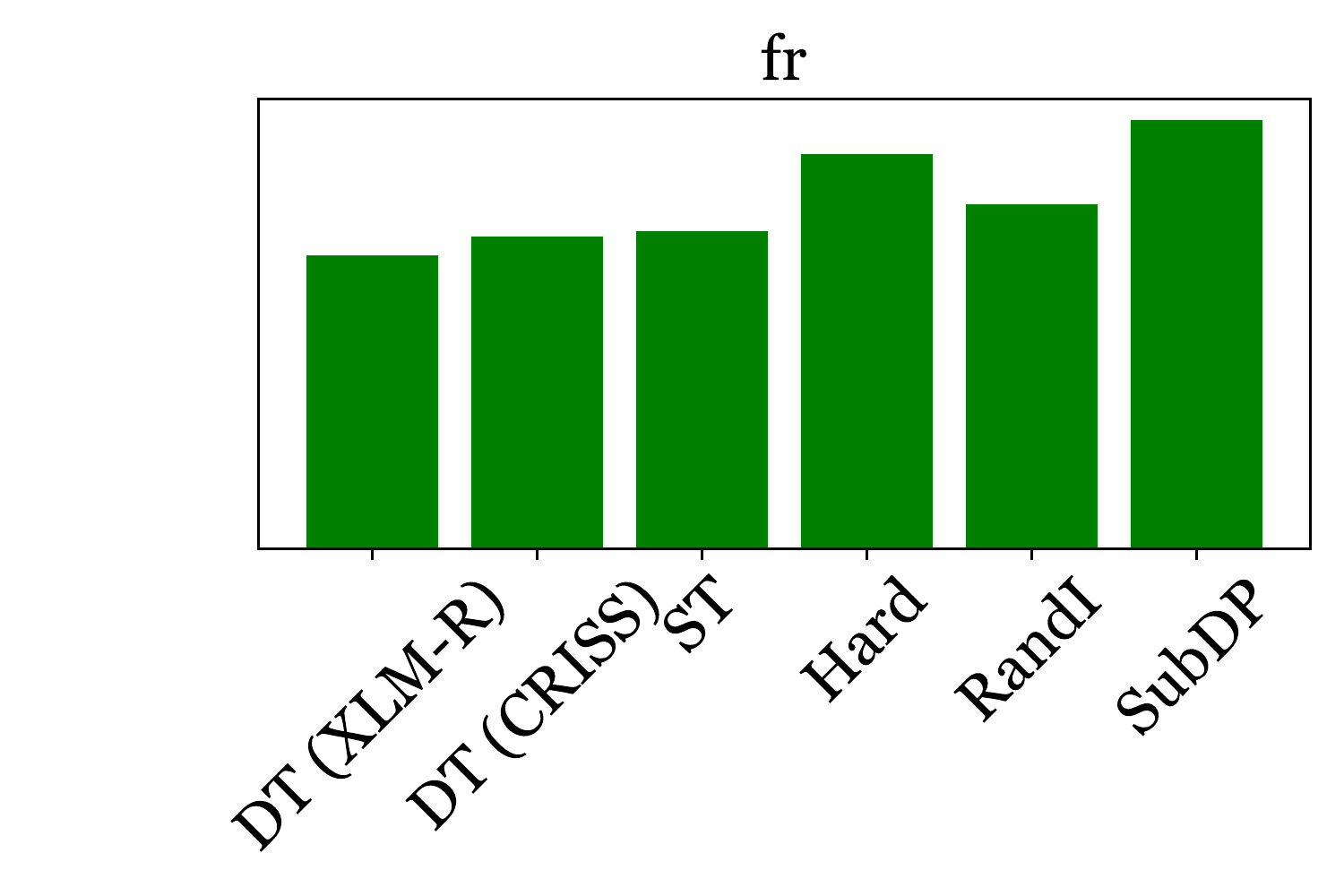} \hspace{-25pt}
\includegraphics[width=0.27\textwidth]{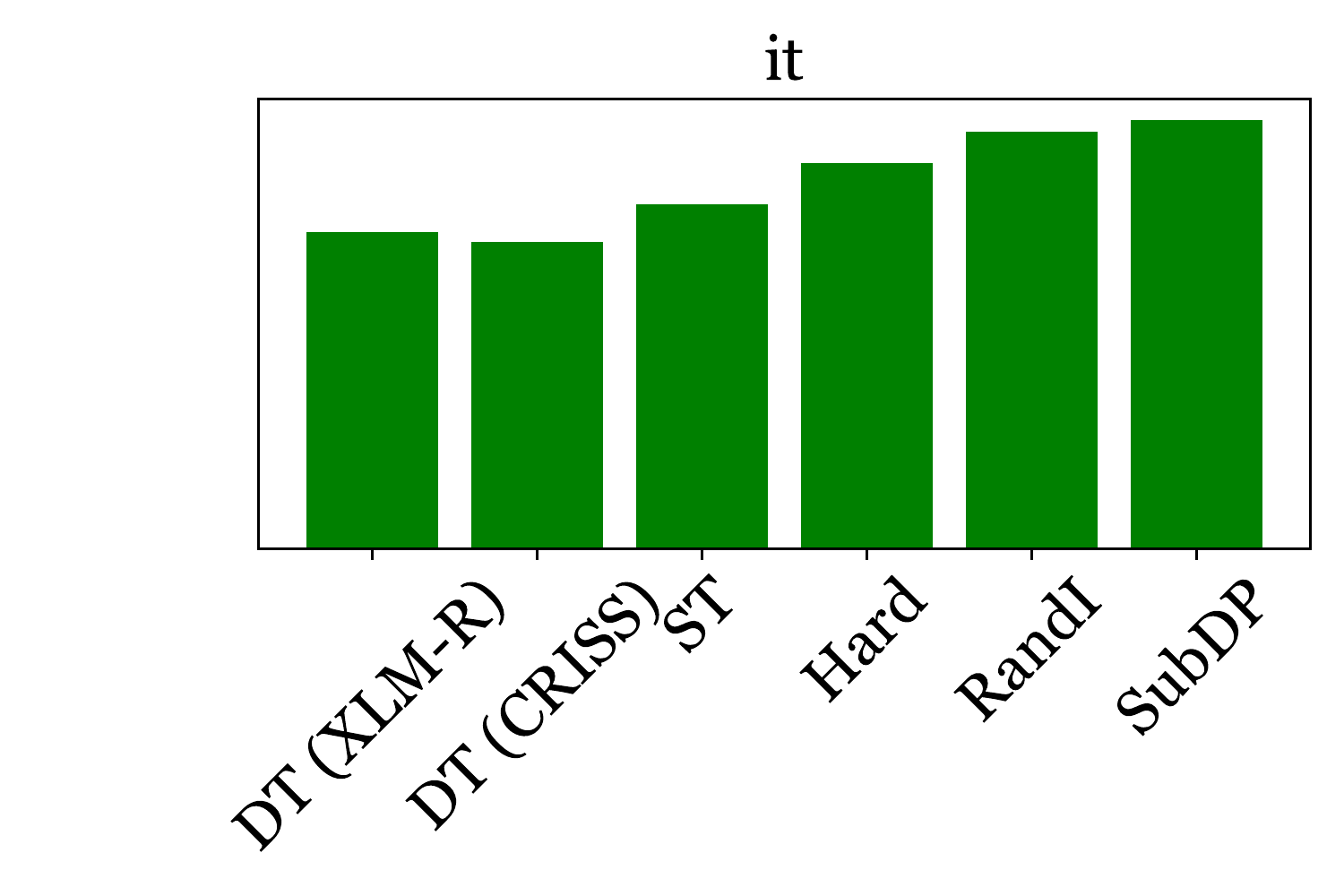} 
\caption{\label{fig:ablation-uas} UAS on the Universal Dependencies v2.2 standard development set. All numbers are averaged across 5 runs. }
\end{figure*}

We present the corresponding UAS results to the LAS in Figure~\ref{fig:ablation} in Figure~\ref{fig:ablation-uas}. 
We arrive at similar conclusions to those reached by LAS trends: \ddp is the only model that consistently ranks among the top contenders and outperforms the direct transfer baseline in all languages.\label{sec:ablation-uas}
\end{CJK*}
\end{document}